\documentclass[runningheads]{llncs}
\usepackage{graphicx}
\usepackage{tikz}
\usepackage{comment}
\usepackage{amsmath,amssymb} % define this before the line numbering.
\usepackage{color}

\usepackage{cases}

\usepackage[accsupp]{axessibility}  % Improves PDF readability for those with disabilities.

\usepackage{bm}

\DeclareMathOperator*{\argmin}{arg\,min}

\def\etal{{et al. }}

\def\ie{\mbox{\textit{i.e.}, }}
\def\eg{\mbox{\textit{e.g.}, }}

\def\mC{{\mathcal C}}

\def\mF{{\mathcal F}}
\def\mG{{\mathcal G}}

\def\mL{{\mathcal L}}

\def\mR{{\mathcal R}}

\def\mT{{\mathcal T}}

\DeclareMathAlphabet\mathbfcal{OMS}{cmsy}{b}{n}
\def\0{{\bf 0}}
\def\1{{\bf 1}}

\def\bI{{\bm{I}}}

\def\bK{{\bm{K}}}

\def\bN{{\bm{N}}}

\def\bR{{\bm{R}}}

\def\bX{{\bm{X}}}
\def\bY{{\bm Y}}
\def\bZ{{\bm{Z}}}

\def\bff{{\bm f}}
\def\bg{{\bm g}}

\def\bs{{\bm s}}

\def\bx{{\bm x}}

\def\bz{{\bm z}}
\def\mmR{{\mathbb R}}

\usepackage{ntheorem}

\newtheorem{thm}{Theorem}
\newtheorem*{*thm}{Theorem}

\newtheorem*{*lemma}{Lemma}

\usepackage{multirow}
\usepackage{hyperref}
\hypersetup{colorlinks=true,urlcolor=blue}

\usepackage{algorithmic}
\usepackage{caption}
\usepackage[linesnumbered,ruled,vlined,onelanguage]{algorithm2e}
\SetKwBlock{DummyBlock}{}{}

\usepackage{wrapfig}
\usepackage{bbding}

\newcommand{\name}{0}
\newcommand{\h}{0}
\newcommand{\w}{0.15}
\newcommand{\wa}{0.15}
\newlength \g
\usepackage{adjustbox}
\newcommand{\etc}{\textit{etc}}
\newcommand{\R}[1]{\textcolor[rgb]{1.00,0.00,0.00}{\textbf{#1}}}
\newcommand{\B}[1]{\textcolor[rgb]{0.00,0.00,1.00}{\textbf{#1}}}

\begin{document}
	% \linenumbers
	\pagestyle{headings}
	\mainmatter
	\def\ECCVSubNumber{3116}  % Insert your submission number here
	
	\title{Towards Interpretable Video Super-Resolution via Alternating Optimization} % Replace with your title Deep Unfolding Network
% 	\title{Space-Time Video Super-Resolution with Interpretable Iterative Optimization} 
	
	\begin{comment}
	\titlerunning{ECCV-22 submission ID \ECCVSubNumber} 
	\authorrunning{ECCV-22 submission ID \ECCVSubNumber} 
	\author{Anonymous ECCV submission}
	\institute{Paper ID \ECCVSubNumber}
	\end{comment}
	%******************
	
% 	\begin{comment}
	\titlerunning{Towards Interpretable Video Super-Resolution via Alternating Optimization}
	\author{Jiezhang Cao\inst{1} \and 
        Jingyun Liang\inst{1} \and
        Kai Zhang\inst{1}\thanks{Corresponding author.} \and
        Wenguan Wang\inst{1} \and
        Qin Wang\inst{1} \and
        Yulun Zhang\inst{1} \and 
        Hao Tang\inst{1} \and 
        Luc Van Gool\inst{1,2}
        }
	\authorrunning{Jiezhang Cao et al.}
	\institute{$^{1}$Computer Vision Lab, ETH Zürich, Switzerland  \quad~~$^{2}$KU Leuven, Belgium \\
    \email{\{jiezhang.cao, jingyun.liang, kai.zhang, wenguan.wang, qin.wang, yulun.zhang, hao.tang, vangool\}@vision.ee.ethz.ch}\\
    % \email{\{jiezcao, jinliang, kaizhang, yawli, yulzhang, wengwang, vangool\}@vision.ee.ethz.ch}\\
    \url{https://github.com/caojiezhang/DAVSR}
    }
% 	\end{comment}
	%******************
	\maketitle
	
	\begin{abstract}
    In this paper, we study a practical space-time video super-resolution (STVSR) problem which aims at generating a high-framerate high-resolution sharp video from a low-framerate low-resolution blurry video. Such problem often occurs when recording a fast dynamic event with a low-framerate and low-resolution camera, and the captured video would suffer from three typical issues: i) motion blur occurs due to object/camera motions during exposure time; ii) motion aliasing is unavoidable when the event temporal frequency exceeds the Nyquist limit of temporal sampling; iii) high-frequency details are lost because of the low spatial sampling rate. These issues can be alleviated by a cascade of three separate sub-tasks, including video deblurring, frame interpolation, and super-resolution, which, however, would fail to capture the spatial and temporal correlations among video sequences. To address this, we propose an interpretable STVSR framework by leveraging both model-based and learning-based methods. Specifically, we formulate STVSR as a joint video deblurring, frame interpolation, and super-resolution problem, and solve it as two sub-problems in an alternate way. For the first sub-problem, we derive an interpretable analytical solution and use it as a Fourier data transform layer. Then, we propose a recurrent video enhancement layer for the second sub-problem to further recover high-frequency details. Extensive experiments demonstrate the superiority of our method in terms of quantitative metrics and visual quality. 
	\end{abstract}

	\section{Introduction}
	Compared with existing space-time video super-resolution (STVSR) methods \cite{xiang2020zooming,xiao2020space}, we mainly focus on the more practical STVSR problem which aims at synthesizing a high space-time resolution (HSTR) clear video from a low space-time resolution (LSTR) blurry video.
    Although great progress has been made in existing STVSR methods, these methods mainly solve the video frame interpolation and video super-resolution tasks jointly, neglecting the motion blur and motion aliasing artifacts~\cite{pollak2020across} that often occur in many real-world scenarios due to limited shutter speed and non-negligible exposure time. Different from exiting methods, we take above two temporal motion degradations into consideration and formulate it as a joint video restoration problem of the video frame interpolation, video deblurring and spatial video super-resolution.
	
\begin{figure}[t]
	\setlength\belowcaptionskip{-10pt}
    \centering
    \begin{minipage}[t]{0.48\textwidth}
    \centering
    \includegraphics[width=6.0cm,trim=0 0 0 0]{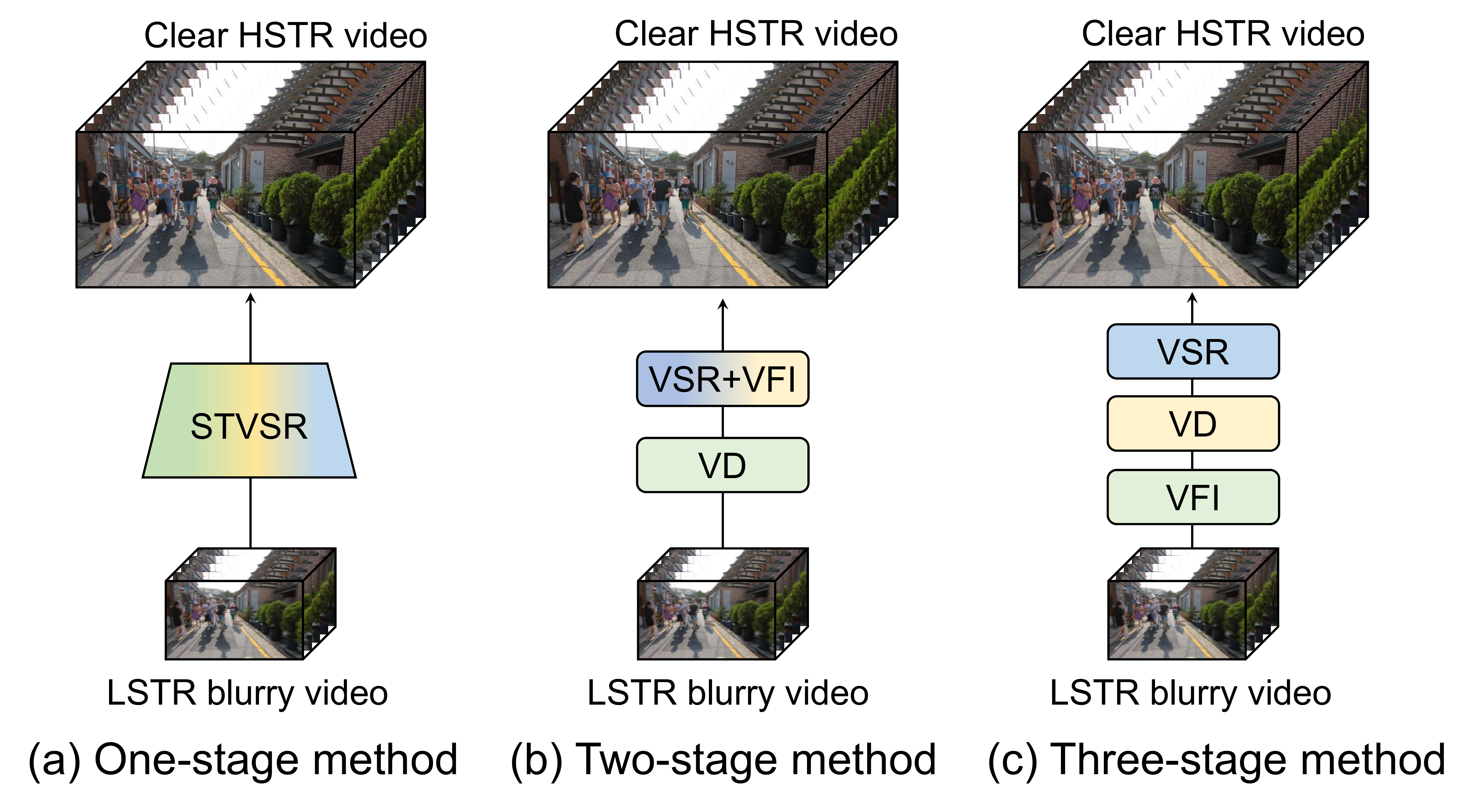}
    \caption{Illustration of one-stage, two-stage and three-stage methods.}
    \label{fig:motivation}
    \end{minipage}
    ~
    \begin{minipage}[t]{0.48\textwidth}
    \centering
    \includegraphics[width=5.8cm,trim=0 30 0 0]{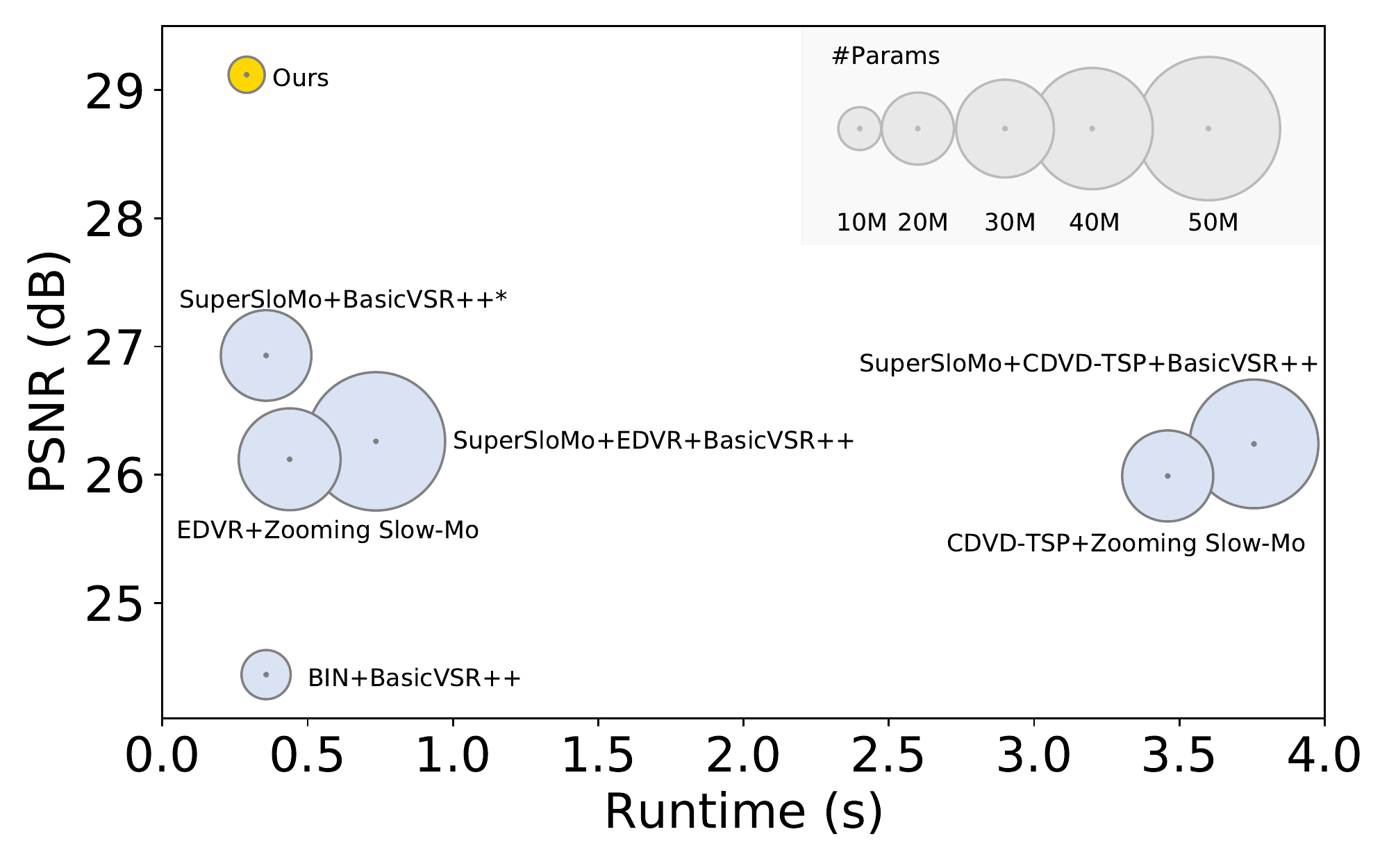}
    \caption{Comparison on performance, runtime and parameter number.}
    \label{fig:para_psnr}
    \end{minipage}
    \end{figure}

	To address the video restoration problem, one straightforward way is to directly combine a video frame interpolation (VFI) method (\eg SuperSloMo \cite{jiang2018superslomo} and DAIN \cite{bao2019dain}), a video deblurring (VD) method (\eg EDVR \cite{wang2019edvr} and CDVD-TSP \cite{pan2020cdvdtsp}), and a video super-resolution (VSR) method (\eg IconVSR \cite{chan2021basicvsr} and BasicVSR++ \cite{chan2021basicvsr++}) in a two-stage or three-stage manner, as shown in Fig. \ref{fig:motivation}. 
	For example, one can firstly interpolate missing intermediate video frames with VFI methods, then deblur the frames with VD methods, and finally super-resolve them with VSR methods.
	In this case, the VFI methods cannot eliminate motion blur nor motion aliasing because they only synthesize new blurry frames, while the VD methods cannot increase the framerate and resolve motion aliasing.

	Alternatively, there are other combinations of VFI, VD and VSR methods, but solving these sub-tasks separately with multi-stage methods may suffer from the following limitations.
	First, ignoring the correlations among VFI, VD and VSR may lead to limited performance since these video restoration tasks are highly intra-related. As will be discussed in Sec.~\ref{sec:vdm}, the degradation from an HSTR clear video to an LSTR blurry video can be well-modelled by a single joint model. A ``divide-and-conquer'' strategy may not be able to benefit from the natural intra-relatedness and suffer from accumulated reconstruction errors from the first stage to the last stage.
	Second, as shown in Fig.~\ref{fig:para_psnr}, the composition of different methods may lead to expensive computational cost and a large number of parameters. This is because the overall runtime and parameter number are the summation of different standalone methods. Therefore, developing a one-stage unified model for the STVSR problem may a better choice.
	
	To solve the above problem, in this paper, we propose a novel STVSR framework that exploits the correlation among different sub-tasks and boosts the overall efficiency significantly. 
	We first reformulate this space-time video super-resolution problem as two sub-problems according to the half quadratic splitting algorithm, and solve them by an analytical solution and a deep learning-based neural network, respectively.  
	More specifically, for the first sub-problem, we solve it based on the fast Fourier transform and propose a Fourier data transform layer to alleviate the motion blur and motion aliasing. 
	For the second sub-problem, we propose a recurrent video enhancement layer with a multi-scale recurrent neural network to enhance the quality of the restored videos.
	Based on the alternating optimization, our end-to-end training method is able to jointly handle video frame interpolation, video deblurring and video super-resolution in STVSR.
	
	The main contributions of this paper are summarized as follows:
	\begin{itemize}
	    \setlength{\itemsep}{4pt}
		\item We formulate a more practical space-time video super-resolution (STVSR) problem by exploring the camera’s intrinsic properties related to motion blur, motion aliasing and other spatial degradation.
		\item We make the first attempt to provide an analytical solution for the STVSR problem by leveraging the model-based methods and learning-based methods. 
		With the help of the analytical solution, we develop a deep alternative video super-resolution network (DAVSR) to improve STVSR performance. 
		\item We propose a new one-stage framework that can address video frame interpolation, video deblurring and video super-resolution simultaneously.
		By exploiting the correlation among the three sub-problems, our method is more effective than two-stage and three-stage methods. 
		\item Our method achieves the state-of-the-art performance on both the REDS4 and Vid4 datasets. It is able to restore high-resolution and high-framerate videos even when the input videos have severe motion blur.
		Moreover, it only has a small number of parameters and has fast inference time.
	\end{itemize}

	\section{Related Work}
	In this section, we discuss the related literature for video frame interpolation (VFI), video deblurring, video super-resolution (VSR) and other joint tasks as they are closely related to our practical STVSR problem.

	\textbf{Video frame interpolation} (VFI) aims to synthesize intermediate frames between adjacent frames of the original frames.
	Recent VFI methods~\cite{long2016learning} propose to learn image matching or take local convolution over two input frames with a learned adaptive convolution kernel.
	Meyer \etal \cite{meyer2015phase} propose a phase-based frame interpolation method which represents motion in the phase shift of individual pixels.
	In addition, flow-based video interpolation methods \cite{bao2019dain,bao2019memc,jiang2018superslomo,liu2017video,niklaus2018context} propose to handle motions by estimating the optical flow. 
	However, directly using VFI cannot reduce motion blur and motion aliasing~\cite{pollak2020across}. This can become an potential issue in our STVSR probelm setup as the input videos are blurry.

	\textbf{Video deblurring} aims at removing the blur artifacts from the input videos.
	Depending on the number of required input frames, there are multiple-frame \cite{su2017deep,jin2018learning,kim2017dynamic,nah2017deep,hyun2015generalized,liang2022vrt,liang2022recurrent} and single-frame \cite{tao2018scale,su2017deep,kupyn2018deblurgan} deblurring methods.
	EDVR \cite{wang2019edvr} restores high-quality deblurred frames by first extracting features of multiple inputs and then conducting feature alignment and fusion. 
	To further exploit the temporal information, some recurrent mechanisms based video deblurring methods have been proposed \cite{hyun2017online,zhang2018adversarial,zamir2017feedback,nah2019recurrent}. 
	Zhou \etal \cite{zhou2019spatio} propose a deblurring network based on filter adaptive convolutional layers.
	Recently, CDVD-TSP \cite{pan2020cdvdtsp}, a CNN-based video deblurring method, approaches the problem by optical flow estimation and latent frame restoration steps.
    To solve the STVSR problem, one can combine video deblurring methods with the traditional STVSR methods.
	
	\textbf{Video super-resolution} reconstructs HR video frames from the corresponding LR frames. 
	There are several VSR methods \cite{caballero2017real,tao2017detail,sajjadi2018frame,wang2018learning,xue2019video,liang2022recurrent} that use optical flow for explicit temporal alignment. 
	Recently, RBPN \cite{haris2019recurrent} combines ideas from single- and multiple-frame SR for VSR, and estimates inter-frame motion to generate SR frames. 
	However, the estimated flow is often inaccurate, resulting in poor performance.  
	To address this, DUF \cite{jo2018deep} synthesizes SR frames by generating dynamic upsampling filters and a residual image based on the local spatio-temporal neighborhood of each pixel without explicit motion estimation.
	TDAN \cite{tian2020tdan} proposes deformable alignment at the feature level without computing optical flow.
	Based on TDAN, EDVR \cite{wang2019edvr} aligns frames at the feature level using deformable convolution networks (DCN) in a coarse-to-fine manner, and proposes an attention module to fuse different frames both temporally and spatially.
	However, most of the above methods are computationally inefficient due to many-to-one frameworks. 
	To ease this, recurrent neural networks (RNN) are adopted in VSR methods \cite{chan2021basicvsr,chan2021basicvsr++} for leveraging temporal information. 
	BasicVSR and its extension (\ie IconVSR) \cite{chan2021basicvsr} propose to improve the performance of feature alignment using bidirectional propagation.
	To improve the performance of BasicVSR, BasicVSR++\cite{chan2021basicvsr++} proposes second-order grid propagation and flow guided deformable alignment.
	To address the STVSR problem, one can combine the above VFI, VD and VSR methods in a three-stage manner.

	\textbf{Space-time video super-resolution} (STVSR) aims at synthesizing a high-resolution slow-motion video from a low-framerate and low-resolution video.
	Shechtman \etal \cite{shechtman2002increasing} are among the first to extend SR methods to the space-time domain. 
	Further STVSR methods based on Markov random field \cite{mudenagudi2010space} and motion assisted steering kernel regression \cite{takeda2010spatiotemporal} are then proposed. 
	In addition, Shahar \etal \cite{shahar2011space} explore the degree of the recurrences within natural videos.
	However, these methods are computationally expensive in practice. 
	To address this, Xiang et al. \cite{xiang2020zooming} propose a one-stage STVSR network to directly learn the mapping from partial LR frames to HR frames.
	Different from our problem setting, traditional STVSR methods ignore the blur degradation from the camera, which can be critical in real-world applications.
	
	\textbf{Video frame interpolation and deblurring}
	are a joint video restoration problem \cite{gupta2020alanet,argaw2021motion,oh2021demfi} that generates high-framerate clear videos from low-framerate blurry inputs.
	Recently, Shen \etal \cite{shen2020bin} propose a pyramid module and an inter-pyramid recurrent module to enhance the restoration quality and exploits the spatio-temporal information, respectively.
	Although these methods deal with motion blur and aliasing, spatial degradation is not considered as in STVSR.
	In addition, Pollak \etal \cite{pollak2020across} propose a deep internal learning approach by exploiting the recurrence within and across different spatio-temporal scales of the video. However, it is a zero-shot temporal-SR and is difficult to improve the SR performance without supervised information.

	\begin{figure}[t]
	\setlength\belowcaptionskip{-5pt}
		\centering
		\includegraphics[width=0.9\linewidth,trim=100 20 0 60]{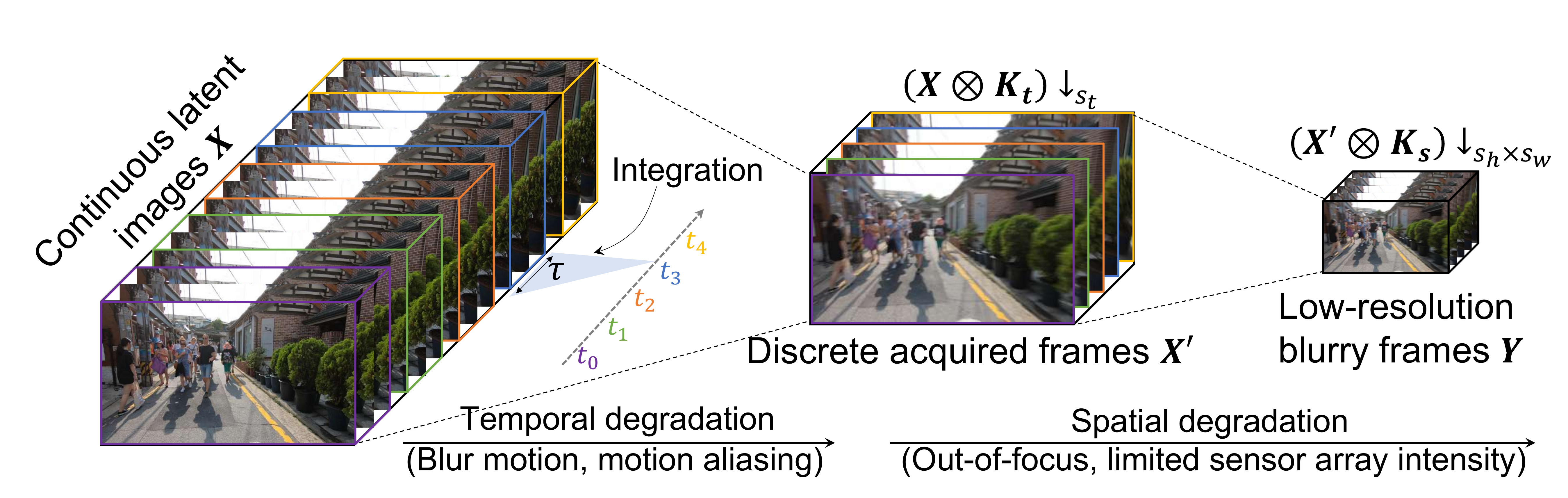}
		\caption{Illustrations of our video degradation. Note that we show temporal and spatial degradation separately for clarity, although they occur simultaneously. Camera sensors capture discrete frames at the time step $t_i (i{\ge} 0)$ by integrating the continuous latent images within an exposure time interval $\tau$, leading to temporal degradation. Then, non-ideal imaging factors such as out-of-focus and limited sensor array intensity result in spatial degradation as well. 
		These two degradations can be implemented by using a temporal kernel $\bK_t$ and a spatial kernel $\bK_s$ with the downsampling $\downarrow_{s_t} $ and $\downarrow_{s_h{\times}s_w}$. }
		\label{fig:degradation}
	\end{figure}

	\section{Proposed Method}
	\subsection{Video Degradation Model}
	\label{sec:vdm}
	We present the video degradation model for the practical STVSR task, as shown in Fig. \ref{fig:degradation}.
	In general, a sequence of video frames is captured by a camera with a periodically on-and-off shutter \cite{telleen2007synthetic,shen2020bin}. When the shutter is open, the camera sensors collect reflected photons and convert them into electrical signals. This can be formulated as an integration of luminous intensity over the exposure time, during which the motion blur may occur if the object moves or the camera shakes. Besides, due to limited shutter on-and-off frequency (framerate), motion aliasing may also occur % become a problem 
	when the temporal dynamic event frequency is beyond the Nyquist limit of framerate. In addition to above temporal degradation, video capturing also suffers from similar spatial degradation to single image capturing as a result of non-ideal imaging factors such as out-of-focus and limited sensor array intensity~\cite{liang21manet}.
	Formally, given a high spatio-temporal resolution (HSTR) video $\bX \in \mmR^{T_h{\times}H_h{\times}W_h{\times}3}$, a 3D blur kernel $\bK$, a low spatio-temporal resolution (LSTR) video $\bY\in \mmR^{T_l{\times}H_l{\times}W_l{\times}3}$ can be formulated as
	\begin{align}
		\bY = (\bX \otimes \bK)\downarrow_{s_t{\times}s_h{\times} s_w} + ~\bN,
		\label{eqn:degradation}
	\end{align}
	where $\otimes$ represents the 3D convolution, and $\downarrow_{s_t{\times}s_h{\times} s_w}$ (abbreviated as $\downarrow_{s}$ in the rest of the paper for clarity) denotes the standard $s$-fold downsampling in three directions: temporal, vertical and horizontal directions. $\bN$ is often assumed to be the additive white Gaussian noise with a noise level of $\sigma$.
    In addition, the sizes of $\bX$ and $\bY$ satisfy $T_h=s_tT_l, H_h=s_hH_l$ and $W_h=s_wW_l$.
	In fact, %our degradation model can be applied in VSR, VFI and VD, \etc.
	some popular video restoration tasks, including spatial VSR, VFI and VD, \etc, can be seen as special cases of the above degradation model. 
	Note that Eq. \eqref{eqn:degradation} can also be used for bicubic degradation since it can be approximated by blur and downsampling with a center shift \cite{zhang2020usrnet}.
	
	\subsection{Problem Setting and Optimization Difficulty}
	\textbf{Problem formulation.}
	The goal of practical STVSR is to solve the joint video restoration tasks, including video frame interpolation, video deblurring and super-resolution.
	Specifically, given a low-framerate and low-resolution (low spatio-temporal resolution) blurry video $\bY$, a 3D blur kernel $\bK$ and downsampling $\downarrow_{s}$, we propose to restore a high-framerate and high-resolution (high spatio-temporal resolution) video $\bX$.
	According to the Maximum A Posteriori (MAP) framework, we solve the problem by minimizing the energy function $E(\bX)$,
	\begin{align}
		\widehat{\bX} = \argmin\limits_{\bX} E(\bX) := \frac{1}{2\sigma^2} \underbrace{\| \bY - (\bX \otimes \bK)\downarrow_{s} \|^2}_{\text{data fidelity term}} + \lambda \!\!\! \underbrace{\Phi(\bX)}_{\text{prior term}}, \label{eqn:energy}
	\end{align}
	where $\lambda$ is a trade-off parameter, the data fidelity term is associated with the model likelihood for reconstruction, and the prior term is a regularization which is related to the prior information of the high spatio-temporal resolution video.
	However, the prior term is often unknown in practice, and thus it is intractable to directly compute an analytical solution to Problem \eqref{eqn:energy}.
	Compared with classic image SR problem, our task is more challenging since the high spatio-temporal resolution video lose high-frequency details in space and time.

	\paragraph{\textbf{\emph{Alternating optimization.}}}
	Based on the Half-Quadratic Splitting (HQS) algorithm \cite{afonso2010fast,zhang2020usrnet}, we introduce an auxiliary variable $\bZ$ that is close to $\bX$, and we have a regularization $\|\bZ - \bX\|^2$ with a penalty parameter $\mu$. 
    Then, we reformulate Problem \eqref{eqn:energy} as the following optimization problem:
	\begin{align}
		E(\bX, \bZ) = \frac{1}{2\sigma^2} \| \bY - (\bZ \otimes \bK)\downarrow_{s} \|^2 + \lambda \Phi(\bX) + \frac{\mu}{2} \| \bZ - \bX\|^2. \label{eqn:new_energy}
	\end{align}
	Then, Problem~\eqref{eqn:new_energy} can be solved by alternately optimizing two sub-problems Eq.~\eqref{eqn:Z} (for $\bZ$) and Eq.~\eqref{eqn:X} (for $\bX$) as follows
	\begin{numcases}{}
		\bZ_k=\argmin_{\bZ} \| \bY - (\bZ \otimes \bK)\downarrow_{s} \|^2 + \mu \sigma^2 \| \bZ - \bX_{k-1}\|^2, \label{eqn:Z}\\
		\bX_k = \argmin_{\bX} \frac{\mu}{2} \| \bZ_k - \bX\|^2 + \lambda \Phi(\bX).  \label{eqn:X}
	\end{numcases}
	With the help of the alternating optimization, we can calculate the closed-form solution to the sub-problem \eqref{eqn:Z}, and solve the sub-problem \eqref{eqn:X} as a video denoising problem. 
	However, directly finding an analytic solution is time-consuming since it requires the inversion of a high dimensional matrix, whose computational complexity is $O(T_h^3W_h^3H_h^3)$.
	One can use simulation-based methods (\eg Markov Chain Monte Carlo \cite{gilavert2014efficient}) to solve the problem, but it would still be computationally expensive for large videos.
	Inspired by existing image super-resolution methods \cite{zhang2020usrnet,chiche2020deep,pan2021deep}, we take two additional challenging video problems, \ie dynamic blur removal and frame interpolation, into consideration, and derive a new analytical solution for the practical STVSR. Note that it is not a trivial issue for these methods to handle video problems because the practical STVSR suffers from more complex degradation such as motion blur and aliasing.

	\subsection{Analytical Solution}
	To solve the sub-problem \eqref{eqn:Z}, we propose to derive a theorem to compute an analytical solution for the practical STVSR problem.
	To develop the theorem, we introduce the Fourier transform to efficiently exploit intrinsic properties of the downsampling and the blur kernel in the frequency domain. 
	\begin{thm}\label{thm:Z} 
	    Let $\mF$ and $\mF^{-1}$ be the fast Fourier transform (FFT) and inverse FFT, and $\overline{\mF}$ be the complex conjugate of $\mF$.
		Assume the blur kernel $\bK$ and the downsampling $\downarrow_{s}$ satisfy some properties \cite{zhao2016fast}.
		Given a video $\bX_{k-1}$ at the $k$-th iteration and a low-resolution video $\bY$, the solution to Eq.~\eqref{eqn:Z} can be computed using the following closed-form expression \footnote{Please see the detailed proof in the supplementary materials.}, \ie
		\begin{align}
			\bZ_k = \mF^{-1} \!\left( \frac{1}{\alpha_k} \!\!\left( \mF(\bR_{k-1}) {-} \overline{\mF(\bK)} \left( \frac{\left( \mF(\bK) \mF(\bR_{k-1}) \right) \downarrow_{s}^{a}  }{\left( s\alpha_k \bI {+} \left(\mF(\bK) \overline{\mF(\bK)} \right) \downarrow_{s}^{a} \right)} \right)\!\!\big\uparrow_{\!{s}}^r \right) \right), \label{eqn:FFT}
		\end{align}
		where $\bR_{k-1}{=}\overline{\mF(\bK)} \mF(\bY\!\uparrow_s)-\alpha_k \bX_{k-1}$ with $\alpha_k{=}\mu_k \sigma^2$, $\uparrow_s$ is a standard s-fold upsampler , \ie upsampling the spatial size by filling the new entries with zeros, $\uparrow_{\!s}^r$ is an upsampler by repeating the tensor the desired dimension, and  $\downarrow_{s}^{a} $ is a distinct block downsampler, \ie averaging the $s_t{\times}s_h{\times}s_w$ distinct blocks. 
	\end{thm}
    In Theorem \ref{thm:Z}, we are able to derive an analytical solution to the sub-problem \eqref{eqn:Z}. 
    Note that the assumptions are not strong and they are widely used in existing studies \cite{robinson2010efficient,vsroubek2011superfast}.
    For example, the assumption of the blur kernel does not depend on the shape and it can be used in different kinds of blurring, such as motion blur and out-of-focus blur \cite{zhao2016fast}. %, atmospheric turbulence
    Different from USRNet \cite{zhang2020usrnet}, our theorem is more general and can be applied to more image or video restoration tasks, including classic image super-resolution, video frame interpolation, video deblurring and the space-time video super-resolution.
    Based on the analytical solution, we are able to further solve the next sub-problem \eqref{eqn:X}.

    \paragraph{\textbf{\emph{Complexity analysis.}}}
	With the help of Theorem \ref{thm:Z}, we further analyze the complexity of calculating the analytical solution \eqref{eqn:FFT} and show that we are able to improve the effectiveness of computing the analytical solution to sub-problem \eqref{eqn:Z}.
	In the theorem, Eq.~\eqref{eqn:FFT} requires three FFT computations and one inverse FFT computation, which are the most expensive parts in the implementation. Considering the computation complexities of FFT and inverse FFT, the computation complexity of Eq.~\eqref{eqn:FFT} is
	$\mathcal{O}(T_h W_h H_h \log (T_h W_h H_h))$,
	which is much smaller than the computation complexity of directly solving Eq.~\eqref{eqn:energy} (\ie $\mathcal{O}(T_h^3W_h^3H_h^3)$) and can be computed efficiently on the modern GPU devices. More analyses can be found in the supplementary materials.
    With the efficient calculation of $\bZ_k$, we can deal with space-time blur (including motion and spatial blur, encoded in $\bK$) and space-time downsampling (including temporal and spatial downsampling, encoded in $\downarrow_{s}$) in a joint and analytical way. The blur is reduced and the details are restored gradually.

	\begin{figure}[t]
		\centering
		\includegraphics[width=1\linewidth]{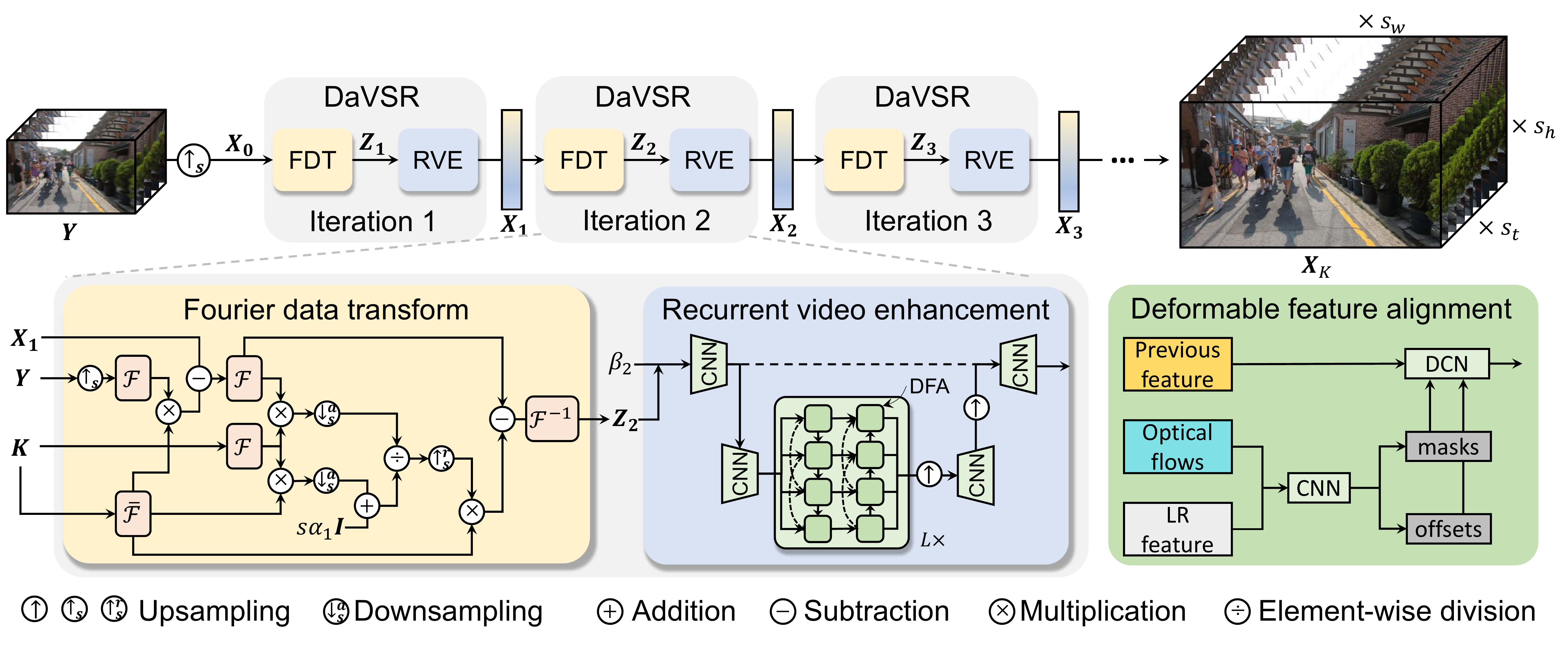}
		\caption{The overall architecture of the proposed method with $K$ iterations. Our one-stage model is able to handle different video degradation (\ie Eq. (\ref{eqn:degradation})) in a joint way and synthesize HSTR frames by taking an LSTR blur video $\bY$, scale factor $s$ and blur kernel $\bK$ as inputs. Specifically, the architecture consists of two main modules, including the FDT layer that reduces the blur degradation, and the RVE layer that makes HR synthesized videos cleaner.}
		\label{fig:framework}
	\end{figure}
	
	\subsection{Deep Alternating Video Super-Resolution Network}
	In this paper, we propose to design a new video super-resolution network based on the alternating optimization, called DAVSR, which solves the sub-problems \eqref{eqn:Z} and \eqref{eqn:X} alternately.
	To this end, we propose a Fourier data transform layer $\mT$ and a recurrent video enhancement layer $\mR$ to address the above two sub-problems, respectively.
	The overall architecture of our method is shown in Fig. \ref{fig:framework}.
	Specifically, the Fourier data transform layer aims to alleviate the video degradation, while the recurrent video enhancement layer aims to enhance synthesized videos by adding more high-frequency details.
	
	\paragraph{\textbf{\emph{Fourier data transform layer.}}}
	With the help of the analytical solution \eqref{eqn:Z}, we aim to reduce the video degradation from the LSTR blurry video.
	During the optimization, we propose to find a clearer HSTR video such that it minimizes a weighted combination of the data fidelity term and the quadratic regularization term.
	To this end, we propose a Fourier data transform (FDT) layer, as shown in Fig. \ref{fig:framework}.
	Specifically, given an LSTR blurry video $\bY$, the scale factor $\bs{=}[s_t, s_h, s_w]$, blur kernel $\bK$ and the parameters $\alpha_k$, we calculate the video by using the analytical solution Eq.~\eqref{eqn:FFT}, \ie
	\begin{align}
		\bZ_k = \mT(\bX_{k-1}|\bY, \bK, \bs, \alpha_k). \label{eqn:data_module}
	\end{align}
	Noted that the Fourier data transform is a model-based method and it has no trainable parameters.
	In this sense, this module has good generalization and it is able to generate meaningful data.
	In addition, this layer is differentiable since every sub-operation is differentiable.
	Compared with USRNet \cite{zhang2020usrnet}, this layer helps simultaneously synthesize high-frequency information in space and time.

	\noindent{\textbf{{Recurrent video enhancement layer.}}}
    For the sub-problem \eqref{eqn:X}, we propose a recurrent video enhancement (RVE) layer to enhance the quality of videos and restore high-frequency sequential textures.
	Such sequential information is important in the continuous video frames, which, however, is neglected in the FDT layer and used in feature alignment.
	To address this, the RVE layer aims to model the sequential dependency and align the features of video frames.
	Specifically, given a video $\bZ_k$ transformed by the FDT layer and the noise level $\beta_k$, the RVE model $\mR$ restores a cleaner HSTR video, \ie
	\begin{align}
		\bX_k = \mR(\bZ_k|\beta_k), \quad\text{where}\;\; \beta_k=\sqrt{{\lambda}/{\mu_k}}.
	\end{align}
	Note that RVE is a learning-based model, and it is implemented by using deformable feature alignment (DFA) module, motivated by \cite{chan2021basicvsr++}.
	Specifically, we first use convolutional layers to extract low-level features $\{{\bg}_k^1, \ldots, {\bg}_k^{T_h}\}$ from the concatenation of $\bZ_k$ and $\beta_k$.
	Then, we use DFA $\mG$ to propagate and align the features $\widetilde{\bz}_{k, j}^i$, and further use residual blocks $\mC$ to fuse the features $\widehat{\bz}_{k, j}^i$, \ie
    \begin{align}
        \widetilde{\bz}_{k, j}^i = \mG \left({\bg}_k^i, \widehat{\bz}_{k, j}^{i-1}, \widehat{\bz}_{k, j}^{i-2}, \bff_k^{i\rightarrow{i{-}1}}, \bff_k^{i\rightarrow{i{-}2}}\right),\\
        \widehat{\bz}_{k, j}^i = \widetilde{\bz}_{k, j}^i + \mC \left( \left[\widehat{\bz}_{k, j-1}^i; \widetilde{\bz}_{k, j}^i \right] \right),~~~~~
    \end{align}
    where $\widehat{\bz}_{k, j}^i$ is the feature at the $i$-th timestep in the $j$-th propagation branch at the $k$-th iteration, $\widehat{\bz}_{k, 0}^i {=} {\bg}_k^i$, and $\bff_k^{i_1\rightarrow{i_2}}$ is the optical flow from $i_1$-th frame to the $i_2$-th frame at the $k$-th iteration, $[\cdot; \cdot]$ is a concatenation along the channel dimension.
    More details of $\mG$ can be found in the supplementary.
    Last, we use convolutional layers to reconstruct $\bx_k^i$ which is the $i$-th element of $\bX_k$.
	
	\IncMargin{1em}
    \begin{algorithm}[t]
    \SetKwInOut{Input}{Input}
    \Input{LSTR video $\bY$, blur kernel $\bK$, scale factor $\bs$, parameters $\alpha_k, \beta_k$}
    {Initialize number of iterations $K$, and initialize $\bX_0$ on $\bY$ in space and time}\;
    \While{\emph{not convergent}}{
    \For{$k\leftarrow 1$ \KwTo $K$}{\label{forins}
    Update $\bZ_k$ by computing $\bZ_k = \mT(\bX_{k-1}|\bY, \bK, \bs, \alpha_k)$\;
    Update $\bX_k$ by computing $\bX_k = \mR(\bZ_k|\beta_k)$\;
    }
    Update the RVE model $\mR$ by minimize the training loss \eqref{loss}.
    }
    \caption{Deep Alternating Video Super-Resolution}\label{alg:vsr}
    \end{algorithm}\DecMargin{1em}

	\paragraph{\textbf{\emph{Loss function and algorithm.}}}
	In the training, our goal is to train the model to minimize the distance between the HSTR video ${\bX}_K$ at the last iteration and the ground-truth video $\bX$. 
	To this end, we use the following Charbonnier loss \cite{charbonnier1994two} because it can handle outliers, \ie
	\begin{align}
		\mL = \sqrt{\| \bX_K - \bX \|^2 + \epsilon^2}, \quad\text{where}\;\; \epsilon=1{\times}10^{-3}.
		\label{loss}
	\end{align}
	Algorithm \ref{alg:vsr} shows the detailed model optimization process. 
	We alternately optimize two sub-problems \eqref{eqn:Z} and \eqref{eqn:X}, and update the RVE model by minimize the loss \eqref{loss}.
	Note that the FDT and RVE modules are differentiable during training.
	At the $K$-th iteration, the algorithm outputs the HSTR video.

	\section{Experiments}
    \noindent{\textbf{Implementation details.}}
	We use Adam optimizer \cite{kingma2014adam} with $\beta_1{=}0.9, \beta_2{=}0.99$, and use Cosine Annealing \cite{loshchilov2016sgdr} to decay the learning rate of the main network from $1{\times}10^{-4}$ to $10^{-7}$.
    In the FAT layer, we use \texttt{torch.fft.fftn} and \texttt{torch.fft.ifftn} to calculate FFT and inverse FFT operators, respectively.
	The total number of iterations for training is 300K, and the number of iterations $K$ for updating $\bZ$ and $\bX$ is 3. 
	The batch size is 4 and the patch size of input LR frames is $64{\times}64$.
	The number of residual blocks for each branch is 7, and the number of feature channels is 64.
%	More experiment details are provided in the supplementary material.

	\noindent\textbf{Datasets and evaluation metrics.}
	REDS \cite{nah2019ntire} is a realistic and dynamic scenes dataset, whose the version of ``blur\_bicubic'' can be used for the practical time-space video super-resolution task.
	The dataset is synthesized by averaging five subsequent frames, and then downsampling these frames with $4\times$ bicubic kernel.
	REDS contains 266 training clips (each with 100 LR frames and 500 GT frames) and 4 testing clips (000, 011, 015 and 020, denoted as REDS4) that have diverse scenes and motions. 
	In addition, another widely used dataset Vid4 \cite{liu2013bayesian} is also used in our experiments. The same degradation model as REDS is used for Vid4 experiments.
    For the evaluation metrics, we use PSNR and SSIM \cite{wang2004image} to measure the performance of VSR methods.

	\subsection{Comparison with State-of-the-art Methods}
	To achieve STVSR, we compare with the following SOTA methods, including single-image SR model (\ie SwinIR \cite{liang2021swinir}), two VFI approaches (\ie Super-SloMo \cite{jiang2018superslomo} and DAIN \cite{bao2019dain}), two deblurring methods (\ie EDVR \cite{wang2019edvr} and CDVD-TSP \cite{pan2020cdvdtsp}), and four recent VSR models, EDVR \cite{wang2019edvr}, BasicVSR \cite{chan2021basicvsr}, IconVSR \cite{chan2021basicvsr}, and BasicVSR++ \cite{chan2021basicvsr++}.
	In addition, we also consider joint video restoration methods, including Zooming Slow-Mo \cite{xiang2020zooming} and BIN \cite{shen2020bin}.

	\begin{table}[!htbp]
		\begin{center}
			\caption{Quantitative comparison of our results and two-stage/three-stage on REDS4. 
			\R{Red} and \B{blue} indicates the best and the second best performance.
			The superscript $^*$ means that the model is trained on the ``blur\_bicubic'' version of REDS, and the superscript $^{\dagger}$ means the pre-trained video deblurring model.}
			\label{table:reds_comparison}
			\resizebox{\textwidth}{!}{
				\begin{tabular}{|l|c|c|c|c|c|c|c|c|c|c|}
					\hline%\noalign{\smallskip}
					\multirow{2}{*}{Methods} & \multicolumn{2}{c|}{Clip\_000} & \multicolumn{2}{c|}{Clip\_011} & \multicolumn{2}{c|}{Clip\_015} & \multicolumn{2}{c|}{Clip\_020} & \multicolumn{2}{c|}{Average} \\
					\cline{2-11}
                    &  PSNR & SSIM & PSNR & SSIM & PSNR & SSIM & PSNR & SSIM & PSNR & SSIM \\
                    \hline
				% 	\noalign{\smallskip}
					\hline
					% \noalign{\smallskip}
					% Bicubic & 24.01&0.631 & 22.87&0.626 & 26.45&0.760 & 21.29&0.605 & 23.66&0.656 \\
					Bicubic+linear interpolation & 23.57&0.606 & 21.75&0.587 & 25.59&0.739 & 20.19&0.564 & 22.78&0.624 \\
					\hline
					BIN+EDVR & 
					25.02&0.709 & 22.00&0.603 & 25.89&0.765 & 21.15&0.572 & 23.52&0.662 \\
					BIN+BasicVSR & 
					25.69&0.716 & 22.49&0.622 & 26.33&0.789 & 21.74&0.583 & 24.06&0.678 \\
					BIN+IconVSR & 
					25.80&0.735 & 22.65&0.632 & 26.52&0.792 & 21.99&0.590 & 24.24&0.687 \\
					BIN+BasicVSR++ & 
					26.01&0.752 & 22.82&0.638 & 26.69&0.799 & 22.23&0.654 & 24.44&0.711 \\
					\hline
					CDVD-TSP+Zooming Slow-Mo & 
					26.74&0.786 & 24.68&0.700 & 29.52&0.859 & 23.02&0.689 & 25.99&0.759 \\
					EDVR+Zooming Slow-Mo & 
					26.89&0.791 & 24.72&0.710 & 29.68&0.861 & 23.19&0.695 & 26.12&0.764 \\
					\hline
					\hline
					DAIN+CDVD-TSP+SwinIR & 
					25.00&0.705 & 24.23&0.689 & 28.22&0.812 & 22.59&0.672 & 25.01&0.720 \\
					DAIN+CDVD-TSP+EDVR & 
					25.95&0.751 & 24.14&0.686 & 28.84&0.839 & 22.61&0.673 & 25.38&0.737 \\
					DAIN+CDVD-TSP+BasicVSR & 26.32&0.777 & 24.23&0.688 & 29.11&0.849 & 22.79&0.681 & 25.61&0.748 \\
					DAIN+CDVD-TSP+IconVSR & 26.40&0.782 & 24.37&0.691 & 29.32&0.855 & 22.86&0.684 & 25.74&0.753 \\
					DAIN+CDVD-TSP+BasicVSR++ & 26.58&0.793 & 24.45&0.691 & 29.57&0.860 & 23.02&0.689 & 25.90&0.758 \\
					\hline
					DAIN+EDVR$^{\dagger}$+SwinIR & 25.56&0.719 & 24.35&0.688 & 28.98&0.839 & 22.83&0.672 & 25.43&0.730 \\
					DAIN+EDVR$^{\dagger}$+EDVR & 26.50&0.774 & 24.25&0.685 & 29.51&0.855 & 22.85&0.674 & 25.78&0.747 \\
					DAIN+EDVR$^{\dagger}$+BasicVSR & 26.66&0.783 & 24.21&0.683 & 29.61&0.858 & 22.76&0.670 & 25.81&0.748 \\
					DAIN+EDVR$^{\dagger}$+IconVSR & 26.75&0.787 & 24.25&0.685 & 29.77&0.863 & 22.80&0.672 & 25.89&0.752 \\
					DAIN+EDVR$^{\dagger}$+BasicVSR++ & 26.91&0.795 & 24.25&0.685 & 29.94&0.868 & 22.82&0.673 & 25.98&0.755 \\
					\hline
					DAIN+EDVR$^*$ & 
					26.29&0.760 & 24.53&0.699 & 29.32&0.848 & 22.96&0.682 & 25.77&0.747 \\
					DAIN+BasicVSR$^*$& 
					26.82&0.790 & 24.71&0.709 & 29.65&0.861 & 23.15&0.693 & 26.08&0.763 \\
					DAIN+IconVSR$^*$ & 
					26.95&0.796 & 24.89&0.713 & 29.89&0.867 & 23.25&0.695 & 26.24&0.768 \\
					DAIN+BasicVSR++$^*$ & 
					27.34&0.814 & 25.11&0.719 & 30.22&0.874 & 23.46&0.703 & 26.53&0.778 \\
					\hline
					\hline
					SuperSloMo+CDVD-TSP+SwinIR & 
					25.26&0.709 & 24.20&0.686 & 28.82&0.835 & 22.76&0.675 & 25.26&0.726 \\
					SuperSloMo+CDVD-TSP+EDVR & 
					26.22&0.753 & 24.18&0.686 & 29.32&0.845 & 22.75&0.674 & 25.62&0.739 \\
					SuperSloMo+CDVD-TSP+BasicVSR & 26.70&0.779 & 24.29&0.689 & 29.57&0.855 & 22.97&0.683 & 25.88&0.752 \\
					SuperSloMo+CDVD-TSP+IconVSR & 26.81&0.785 & 24.42&0.691 & 29.85&0.862 & 23.06&0.686 & 26.04&0.756 \\
					SuperSloMo+CDVD-TSP+BasicVSR++ & 27.06&0.797 & 24.53&0.692 & 30.12&0.868 & 23.24&0.691 & 26.24&0.762 \\
					\hline
					SuperSloMo+EDVR$^{\dagger}$+SwinIR & 25.86&0.726 & 24.32&0.687 & 29.36&0.844 & 22.87&0.672 & 25.60&0.732 \\
					SuperSloMo+EDVR$^{\dagger}$+EDVR & 26.98&0.784 & 24.27&0.685 & 30.04&0.862 & 22.92&0.674 & 26.05&0.751 \\
					SuperSloMo+EDVR$^{\dagger}$+BasicVSR & 27.12&0.793 & 24.24&0.683 & 30.07&0.864 & 22.82&0.670 & 26.06&0.753 \\
					SuperSloMo+EDVR$^{\dagger}$+IconVSR & 27.24&0.798 & 24.27&0.685 & 30.28&0.870 & 22.86&0.672 & 26.16&0.756 \\
					SuperSloMo+EDVR$^{\dagger}$+BasicVSR++ & 27.41&0.805 & 24.28&0.685 & 30.46&0.874 & 22.88&0.673 & 26.26&0.759 \\
					\hline
					SuperSloMo+EDVR$^*$ & 26.66&0.769 & 24.60&0.700 & 29.83&0.855 & 23.11&0.684 & 26.05&0.752 \\
					SuperSloMo+BasicVSR$^*$ & 27.35&0.803 & 24.78&0.710  & 30.16&0.868 & 23.35&0.696 & 26.41&0.769 \\
					SuperSloMo+IconVSR$^*$ & \B{27.52}&0.810 & 24.95&0.714 & 30.48&0.876 & 23.46&0.699 & 26.60&0.775 \\
					SuperSloMo+BasicVSR++$^*$ & \R{28.00}&\R{0.829} & \B{25.19}&\B{0.720} & \B{30.83}&\B{0.883} & \B{23.70}&\B{0.707} & \B{26.93}&\B{0.785} \\
					\hline\hline
					\bf{DAVSR (Ours)}   & 
					27.23&\B{0.820} & \R{30.15}&\R{0.865} & \R{31.05}&\R{0.894} & \R{28.06}&\R{0.858} &
					\R{29.12}&\R{0.859} \\
					\hline
				\end{tabular}
			}
		\end{center}
	\end{table}

	\begin{table}[!htbp]
	\setlength\belowcaptionskip{-20pt}
		\begin{center}
			\caption{Quantitative comparison of our results and two-stage/three-stage methods on Vid4. \R{Red} and \B{blue} indicates the best and the second best performance.
			}
		   	\label{table:vide4_comparison}
			\resizebox{1\textwidth}{!}{
				\begin{tabular}{|l|c|c|c|c|c|c|c|c|c|c|}
					\hline
				% 	\noalign{\smallskip}
					\multirow{2}{*}{Methods} & \multicolumn{2}{c|}{Calendar} & \multicolumn{2}{c|}{City} & \multicolumn{2}{c|}{Foliage} & \multicolumn{2}{c|}{Walk} & \multicolumn{2}{c|}{Average} \\
					\cline{2-11}
                    &  PSNR & SSIM & PSNR & SSIM & PSNR & SSIM & PSNR & SSIM & PSNR & SSIM \\
				% 	\noalign{\smallskip}
					\hline\hline
					DAIN+CDCD-TSP+BasicVSR++ & 17.05&0.480 & 21.64&0.421 & 19.01&0.340 & 19.68&0.668 & 19.35&0.477 \\
					DAIN+EDVR$^{\dagger}$+BasicVSR++ & 17.35&0.497 & 21.80&0.433 & 19.22&0.350 & 19.91&0.687 & 19.57&0.492 \\
					DAIN+BasicVSR++$^*$ & 18.05&0.522 & \B{22.22}&\B{0.439} & 20.08&\B{0.400} & \B{20.33}&\B{0.692} & 20.17&0.513 \\
					\hline
					\hline
					SuperSloMo+CDCD-TSP+BasicVSR++ & 17.11&0.488 & 21.70&0.427 & 19.09&0.344 & 19.78&0.672 & 19.42&0.483 \\
					SuperSloMo+EDVR$^{\dagger}$+BasicVSR++ & 17.59&0.505 & 21.78&0.432 & 19.22&0.346 & 19.86&0.687 & 19.61&0.493 \\
					SuperSloMo+BasicVSR++$^*$ & \B{18.30}&\B{0.530} & {22.21}&{0.439} & \B{20.09}&0.398 & {20.30}&{0.692} & \B{20.22}&\B{0.515} \\
					\hline\hline
					\bf{DAVSR (Ours)}   & 
					\R{22.09}&\R{0.747} & \R{25.18}&\R{0.731} & \R{23.96}&\R{0.680} & \R{24.18}&\R{0.827} &
					\R{23.85}&\R{0.746} \\
					\hline
				\end{tabular}
			}
		\end{center}
	\end{table}
	
	\paragraph{\textbf{\emph{Quantitative comparison.}}}
	From Tables \ref{table:reds_comparison} and \ref{table:vide4_comparison}, we have the following observations. 
	(1) Our method outperforms the two-stage and three-stage methods by a large margin on REDS4 and Vid4.
	Especially for fast motion videos (\eg clips 011 and 020), our method has the more significant improvements than these methods.
	It suggests that our one-stage network is able to exploit diverse spatio-temporal patterns and simultaneously handle VFI, VD and VSR. 
	(2) The SOTA image SR method (\ie SwinIR \cite{liang2021swinir}) performs worse than other VSR methods since it cannot handle sequential information of videos and cannot remove the motion blur.
	(3) The two-stage framework (\eg SuperSloMo+BasicVSR++$^*$) is better than the three-stage framework because the reconstruction error propagates severely along the stages.
% 	For example, DAIN+BasicVSR++$^*$ has the better performance than the compared three-stage approaches. 
	(4) The performance of two-stage and three-stage methods are influenced by different motions.
	REDS4 has diverse scenes and motions, \eg Clip\_000 has small motions and other video clips has more complex motions.
	Moreover, the VFI methods are sensitive to large motions.
	Thus, the performance of SuperSloMo+BasicVSR++$^*$ is better than that of our method on Clip\_000, but is worse on other clips due to large motions.

	\begin{figure*}[!t]  %htbp
        \setlength{\belowcaptionskip}{-0.6cm}
		\centering
		\scriptsize
		%%%%%
		\renewcommand{\h}{0.105}
		\renewcommand{\wa}{0.12}
		\newcommand{\wb}{0.16}
		\renewcommand{\g}{-0.7mm}
		\renewcommand{\tabcolsep}{1.8pt}
		\renewcommand{\arraystretch}{1}
		\resizebox{1\linewidth}{!} {
			\begin{tabular}{cc}
				\renewcommand{\name}{figures/REDS4/00000116_}
				\renewcommand{\h}{0.12}
				\renewcommand{\w}{0.2}
				\begin{tabular}{cc}
					\begin{adjustbox}{valign=t}
						%\tiny
						\begin{tabular}{c}%left bottom right top
							\includegraphics[trim={600 0 0 200},clip, width=0.354\textwidth]{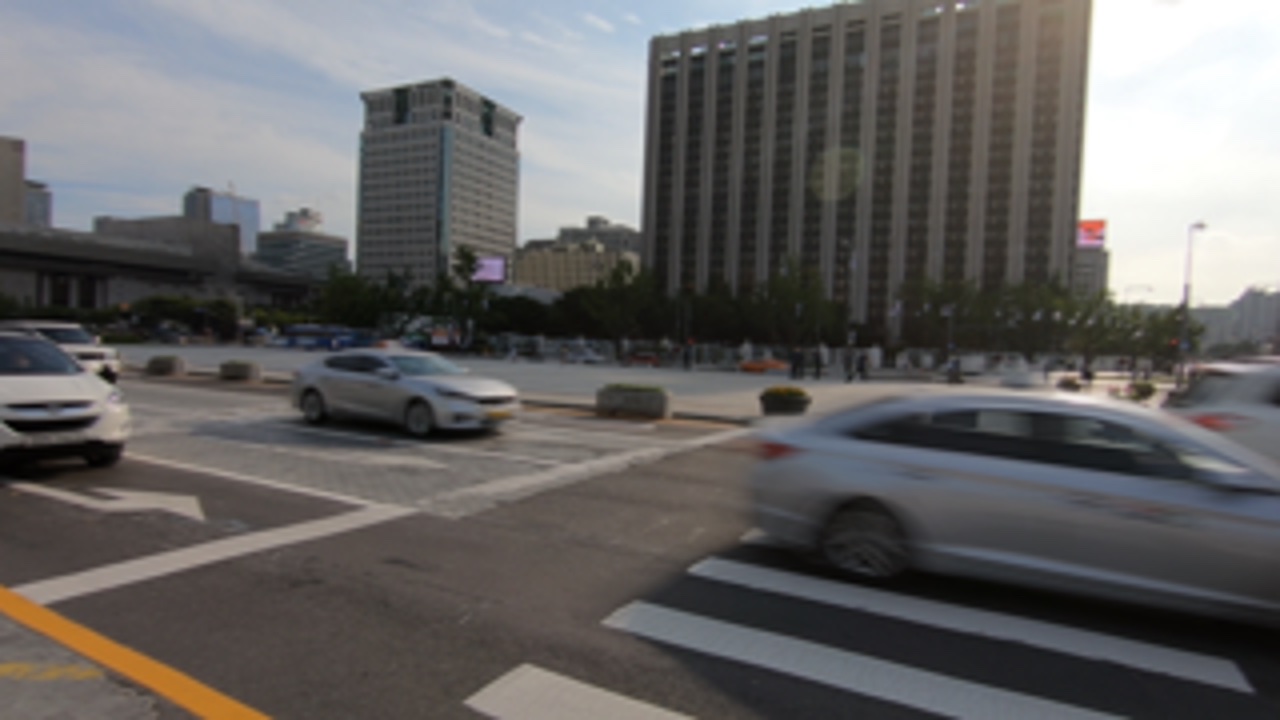}
							\\
							\tiny{Frame 116, Clip 015, REDS4}~\cite{nah2019ntire} 
						\end{tabular}
					\end{adjustbox}
					\begin{adjustbox}{valign=t}
						%\tiny
						\begin{tabular}{cccccc}
							\includegraphics[trim={950 300 200 320},clip,height=\h \textwidth, width=\w \textwidth]{\name bicubic.jpeg} &
							\includegraphics[trim={950 300 200 320},clip,height=\h \textwidth, width=\w \textwidth]{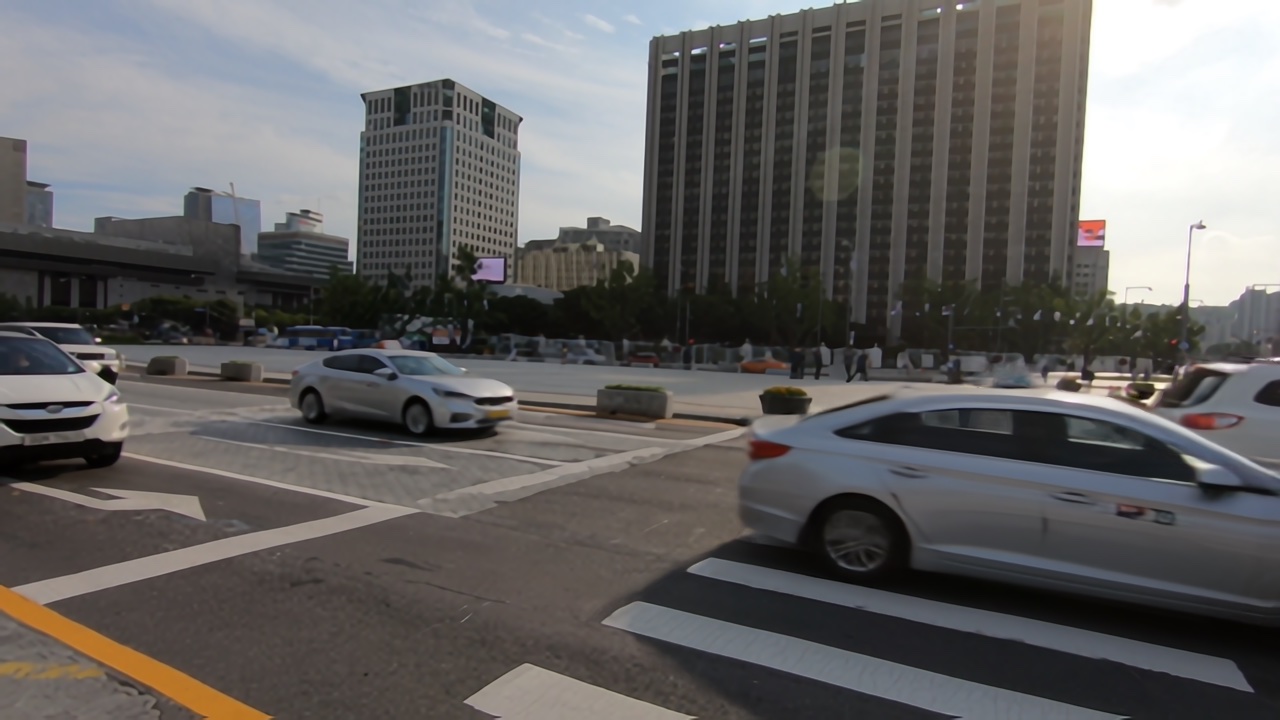} &
							\includegraphics[trim={950 300 200 320},clip,height=\h \textwidth, width=\w \textwidth]{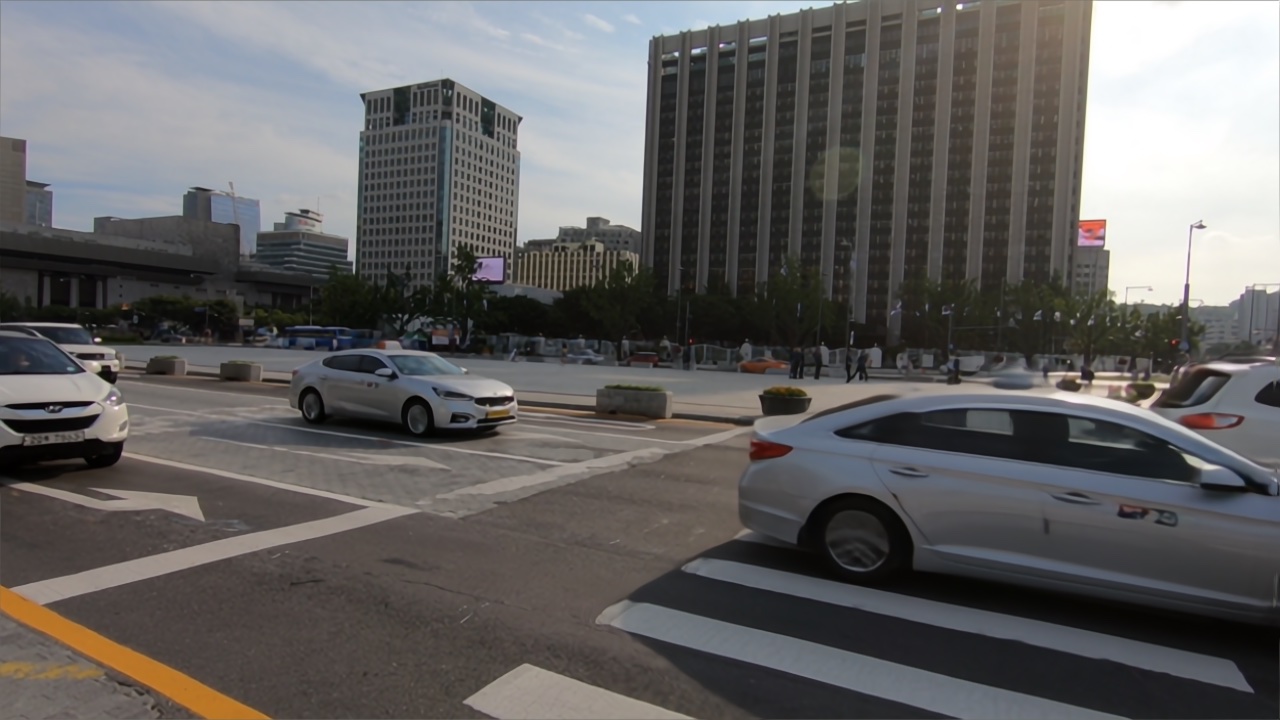} &
							\includegraphics[trim={950 300 200 320},clip,height=\h \textwidth, width=\w \textwidth]{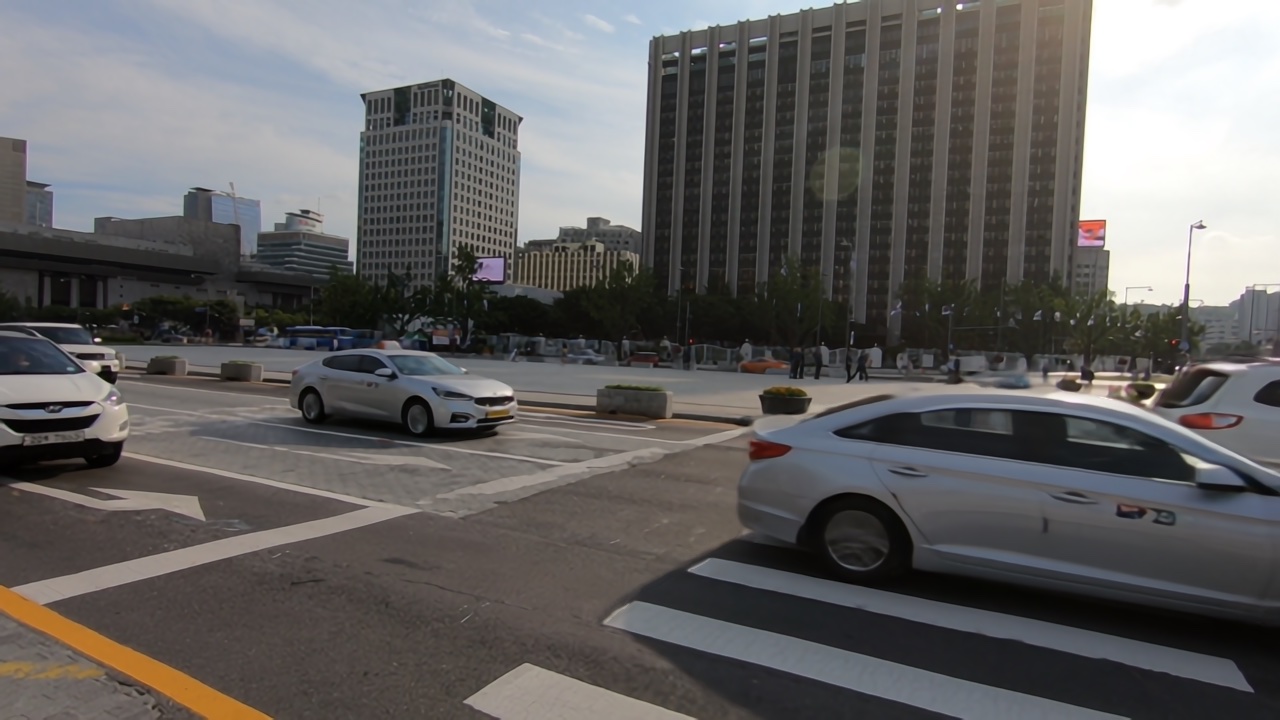} 
							\\
							\tiny{Bicubic} &
							\tiny{DAIN$\!$+EDVR$^{\dagger}$\!+SwinIR} & \!\!\!\tiny{SloMo\!+BVSR}\! &
							\tiny{DAIN+BVSR} 
							\\
							\includegraphics[trim={950 300 200 320},clip,height=\h \textwidth, width=\w \textwidth]{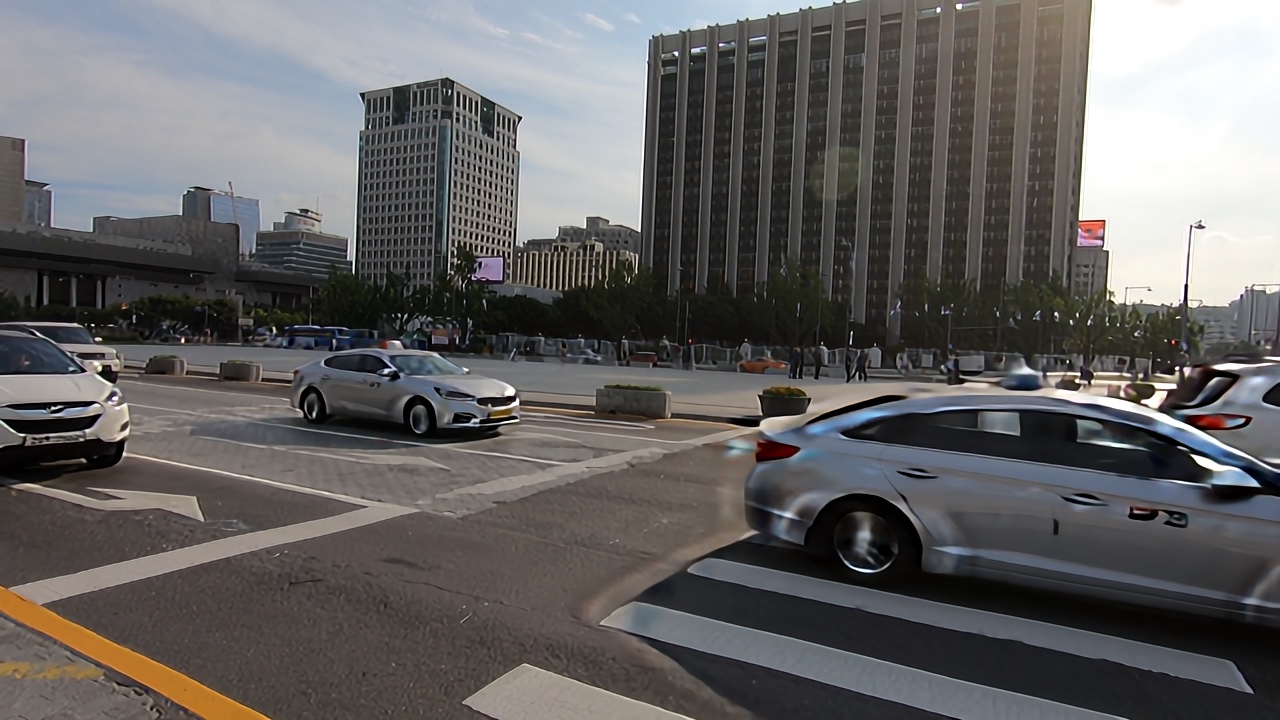} &
							\includegraphics[trim={950 300 200 320},clip,height=\h \textwidth, width=\w \textwidth]{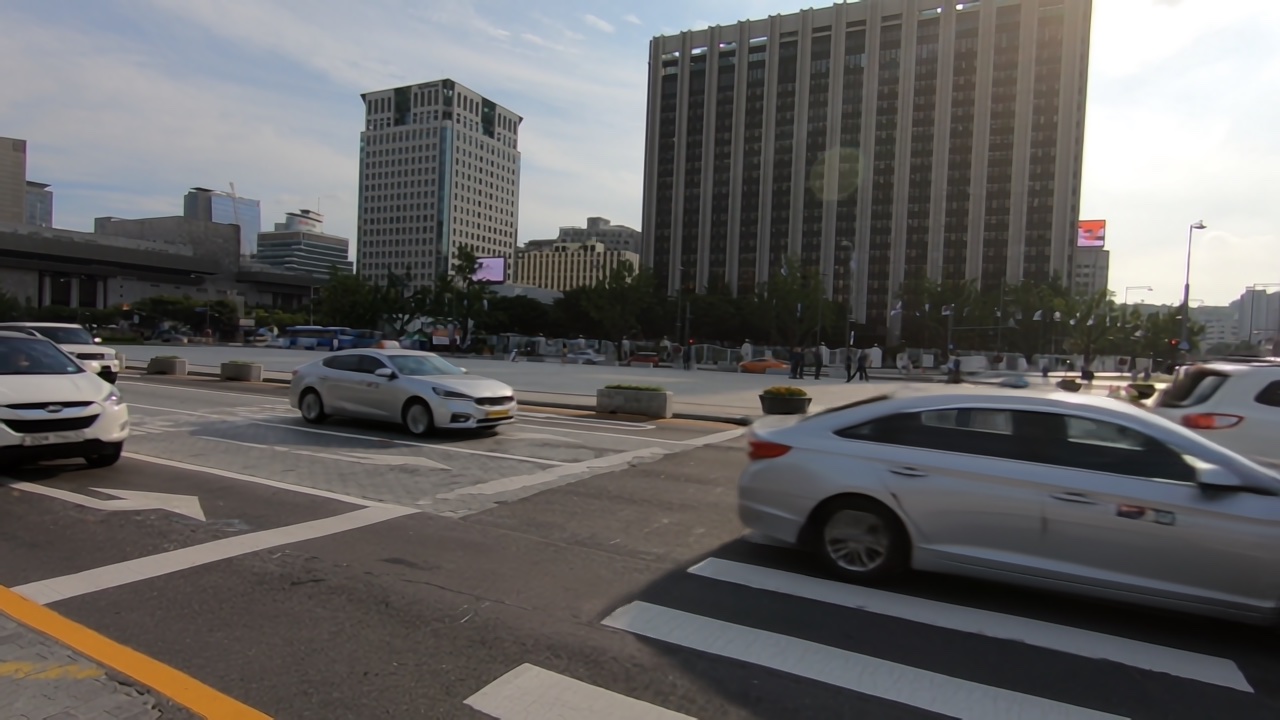} &
							\includegraphics[trim={950 300 200 320},clip,height=\h \textwidth, width=\w \textwidth]{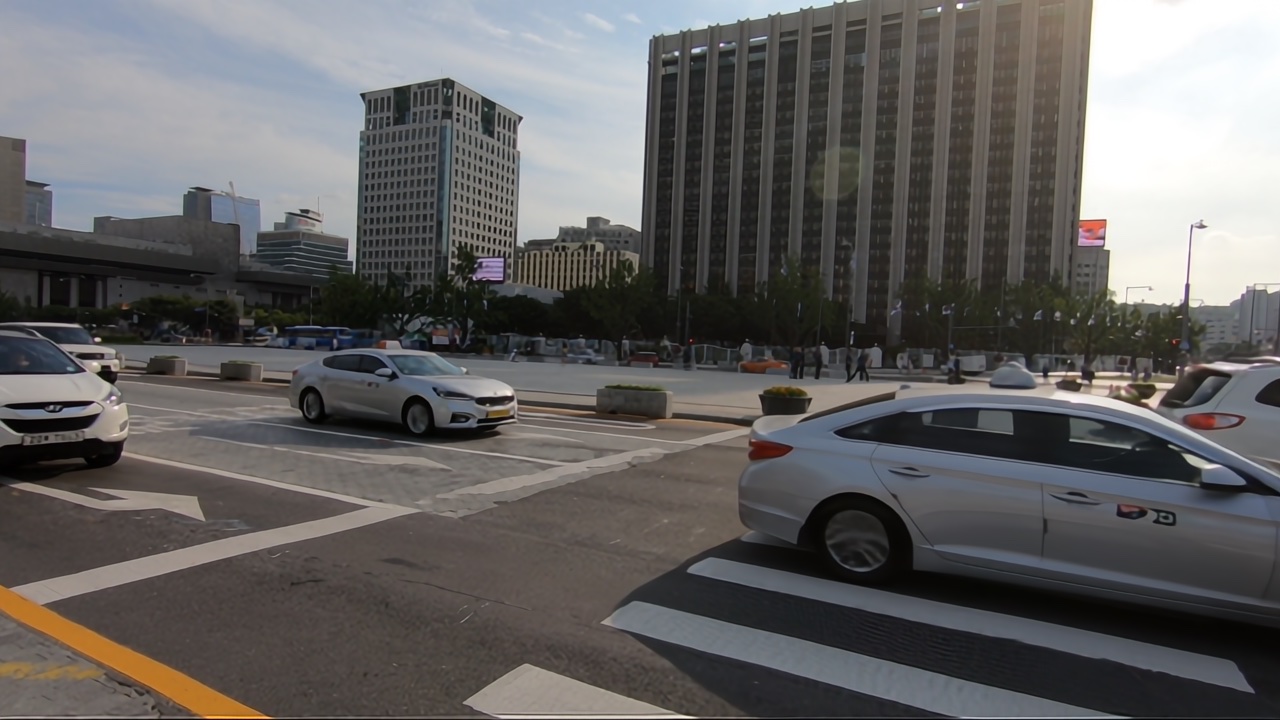}
							&		
							\includegraphics[trim={950 300 200 320},clip,height=\h \textwidth, width=\w \textwidth]{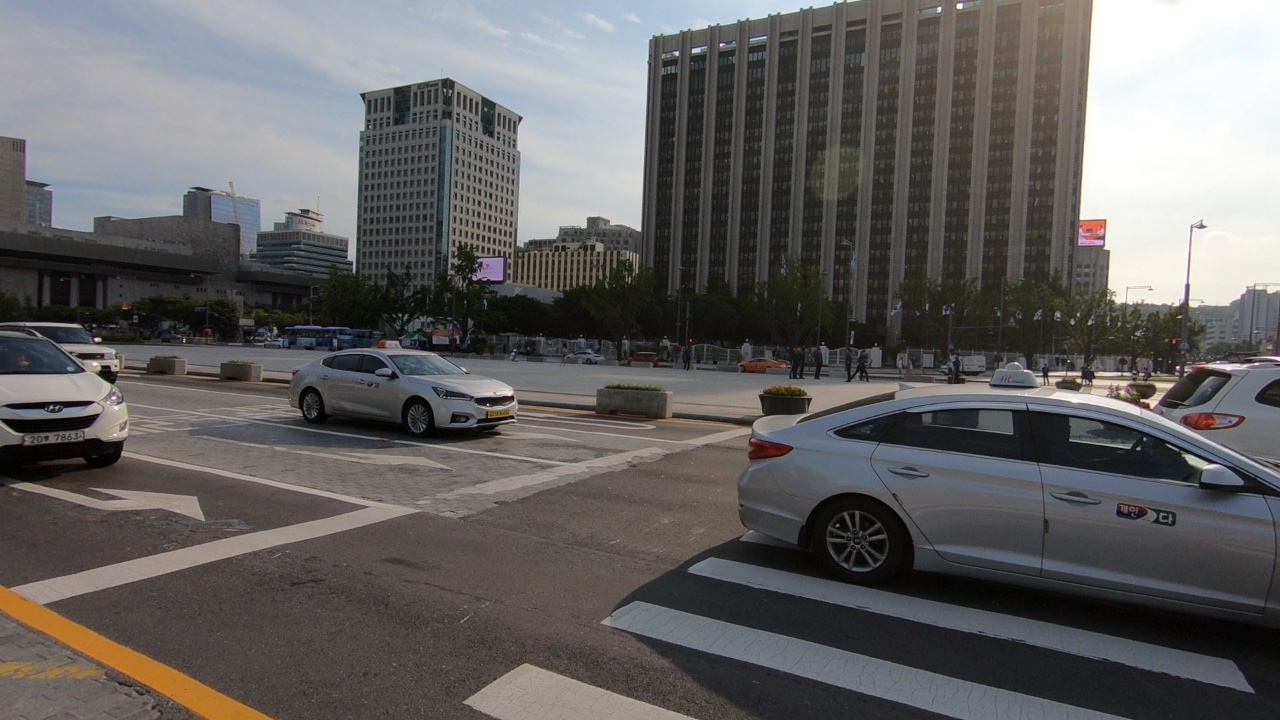} 
							\\ 
							\tiny{BIN+BVSR} &	\!\!\!\tiny{DAIN\!+EDVR\!+BVSR}\!\!\!\!\!\!   &
							\tiny{\textbf{Ours}} &
							\tiny{GT}
							\\
						\end{tabular}
					\end{adjustbox}
				\end{tabular}
			\end{tabular}
		}
		\\ 
		\resizebox{1\linewidth}{!} {
			\begin{tabular}{cc}
				
				\renewcommand{\name}{figures/REDS4/00000040_}
				\renewcommand{\h}{0.12}
				\renewcommand{\w}{0.2}
				\begin{tabular}{cc}
					\begin{adjustbox}{valign=t}
						%\tiny
						\begin{tabular}{c}%left bottom right top
							\includegraphics[trim={0 10 380 20},clip, width=0.354\textwidth]{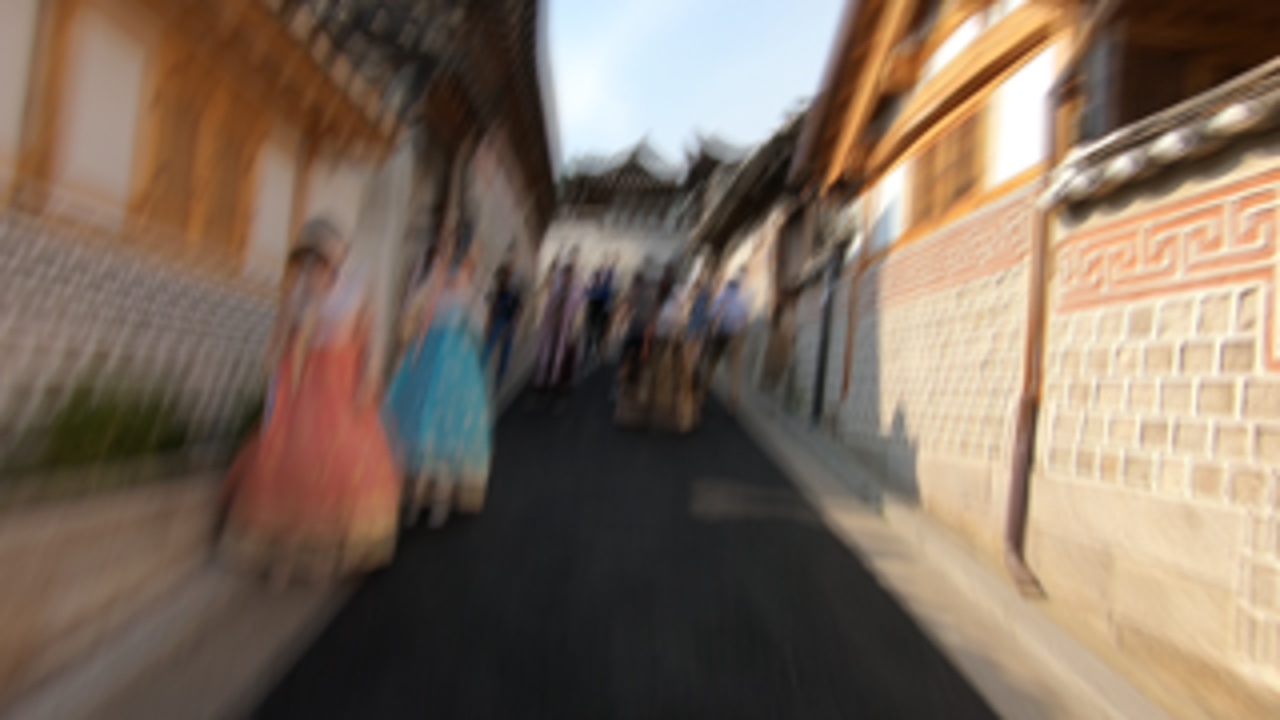}
							\\
							\tiny{Frame 040, Clip 011, REDS4~\cite{nah2019ntire}}
						\end{tabular}
					\end{adjustbox}
					\begin{adjustbox}{valign=t}
						%\tiny
						\begin{tabular}{cccccc}
							\includegraphics[trim={0 400 1220 200},clip,height=\h \textwidth, width=\w \textwidth]{\name bicubic.jpeg} &
							\includegraphics[trim={0 400 1220 200},clip,height=\h \textwidth, width=\w \textwidth]{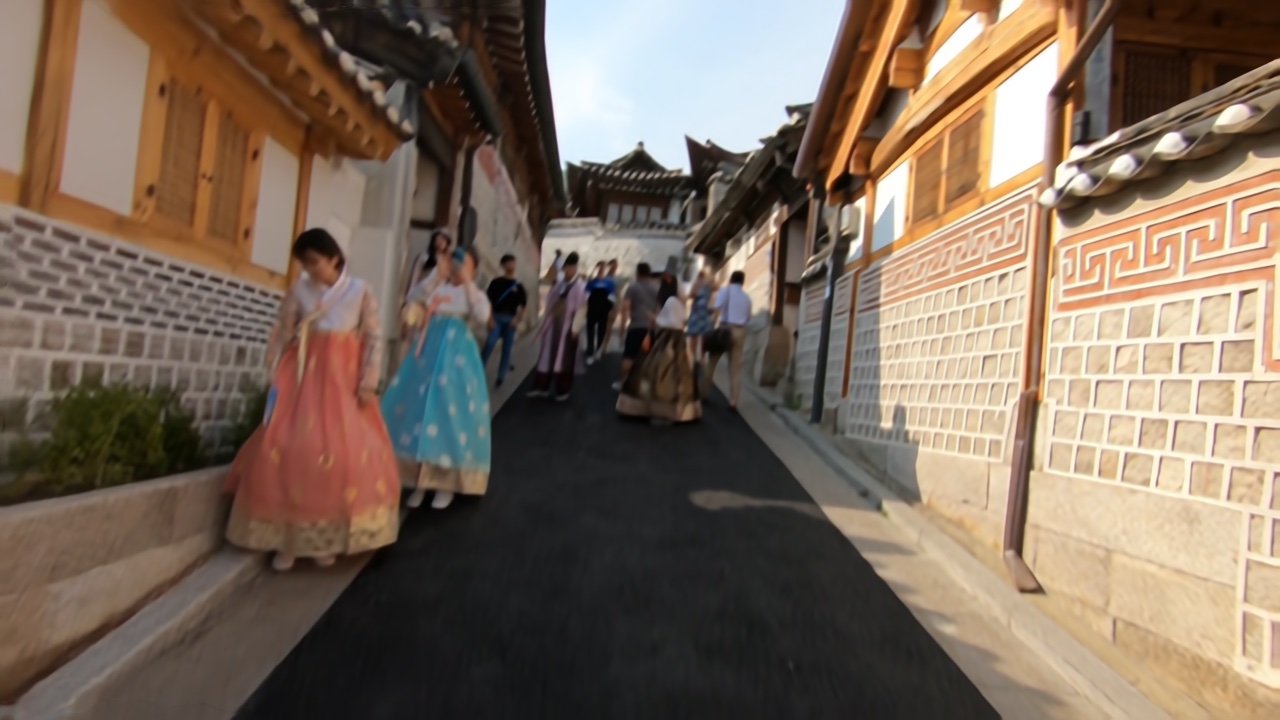} &
							\includegraphics[trim={0 400 1220 200},clip,height=\h \textwidth, width=\w \textwidth]{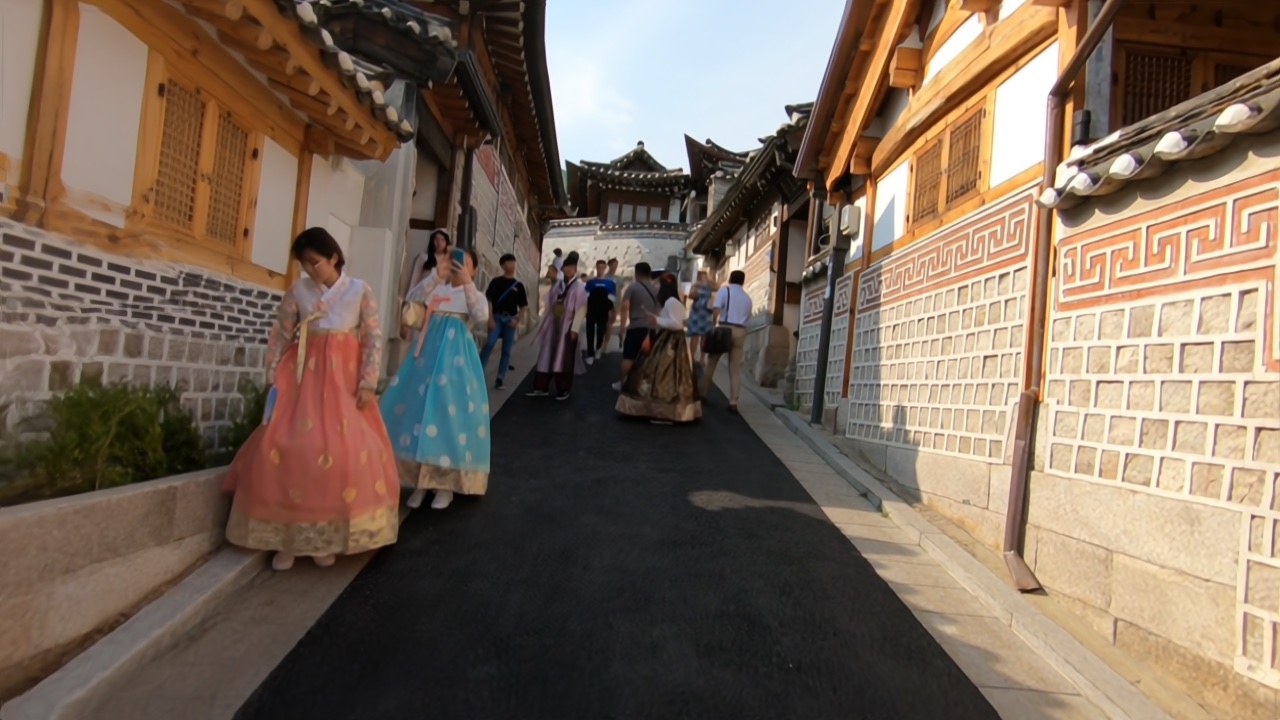} &
							\includegraphics[trim={0 400 1220 200},clip,height=\h \textwidth, width=\w \textwidth]{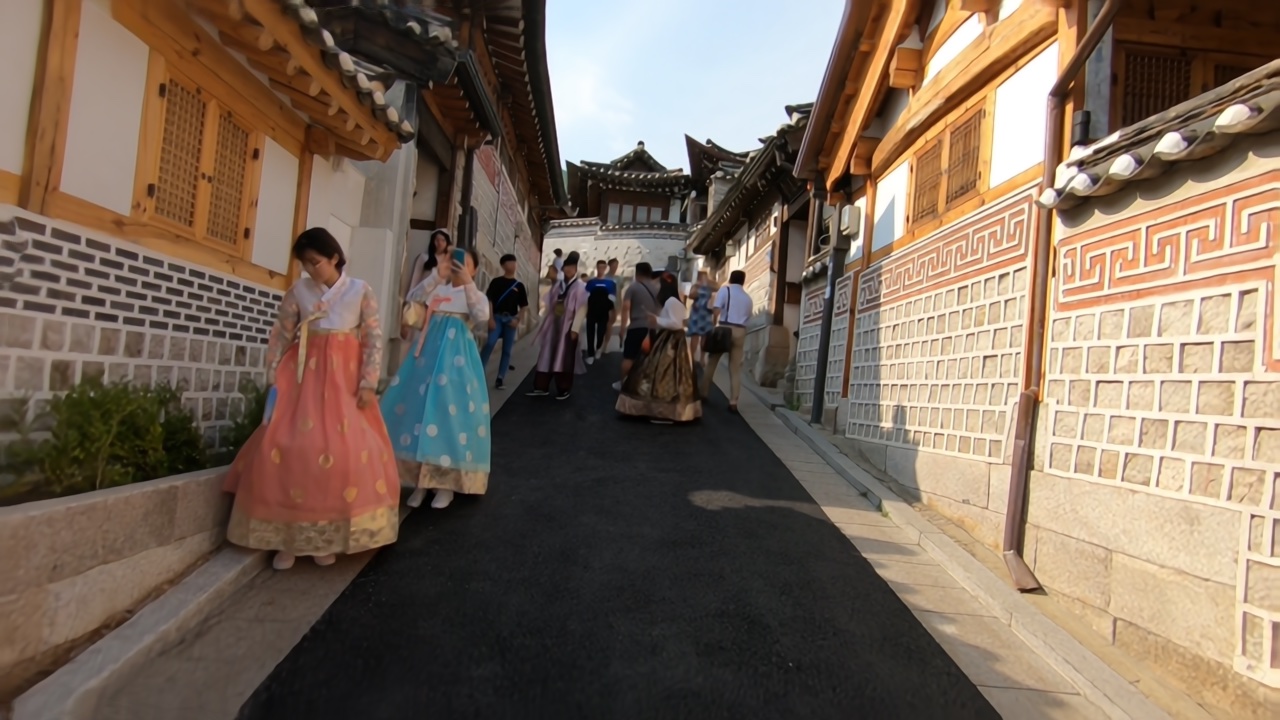} 
							\\
							\tiny{Bicubic} &
							\tiny{DAIN$\!$+EDVR$^{\dagger}$\!+SwinIR} & \!\!\!\tiny{SloMo$\!$+BVSR}\! &
							\tiny{DAIN+BVSR} 
							\\
							\includegraphics[trim={0 400 1220 200},clip,height=\h \textwidth, width=\w \textwidth]{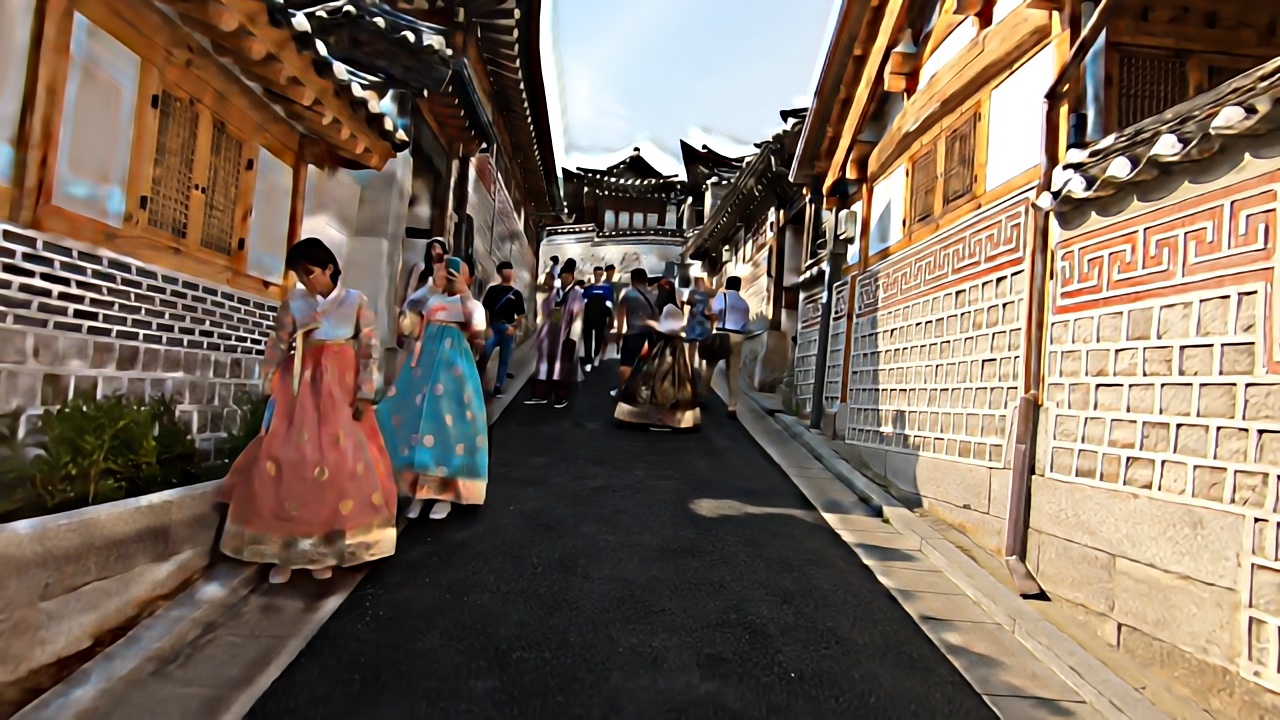} &
							\includegraphics[trim={0 400 1220 200},clip,height=\h \textwidth, width=\w \textwidth]{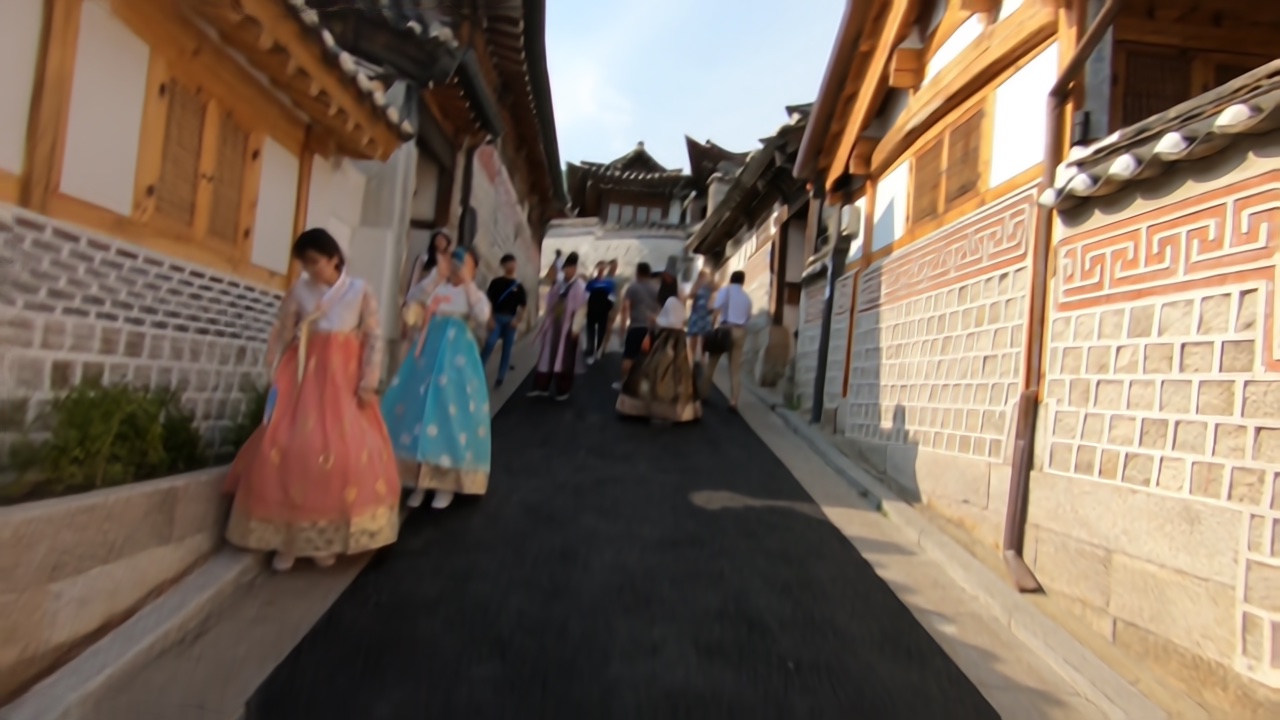} &
							\includegraphics[trim={0 400 1220  200},clip,height=\h \textwidth, width=\w \textwidth]{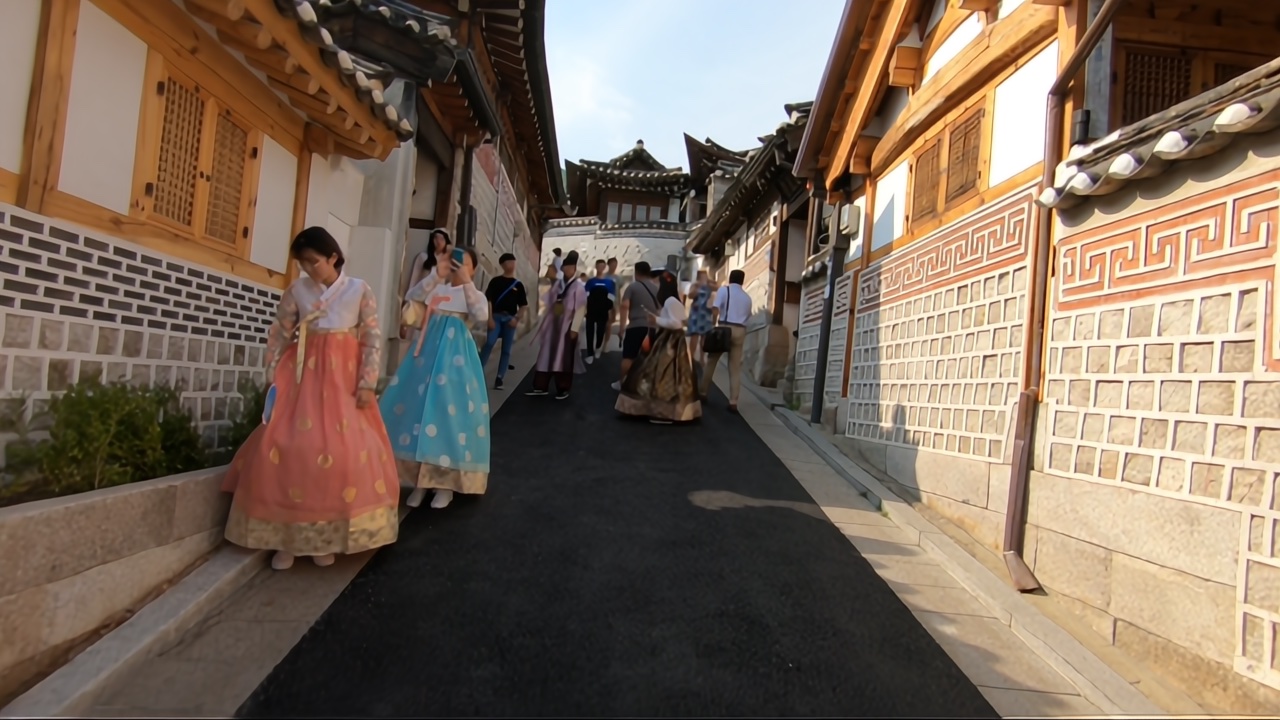} &		
							\includegraphics[trim={0 400 1220 200},clip,height=\h \textwidth, width=\w \textwidth]{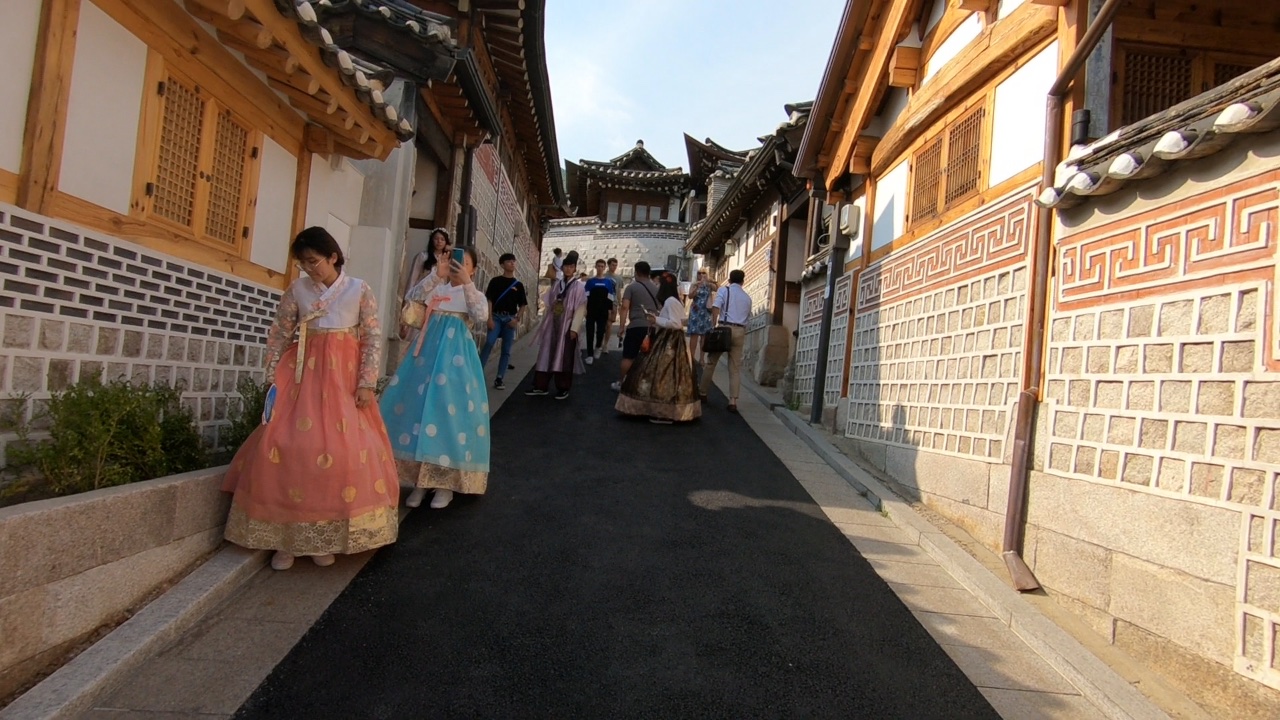} 
							\\ 
							\tiny{BIN+BVSR} &	\tiny{DAIN\!+EDVR$^{\dagger}$\!+BVSR} &
							\tiny{\textbf{Ours}} &
							\tiny{GT}
							\\
						\end{tabular}
					\end{adjustbox}
				\end{tabular}
			\end{tabular}
		}
        % \\
        \resizebox{1\linewidth}{!} {
			\begin{tabular}{cc}
				\renewcommand{\name}{figures/Vid4/00000030_}
				\renewcommand{\h}{0.12}
				\renewcommand{\w}{0.2}
				\begin{tabular}{cc}
					\begin{adjustbox}{valign=t}
						%\tiny
						\begin{tabular}{c}%left bottom right top
							\includegraphics[trim={40 10 0 50},clip, width=0.354\textwidth]{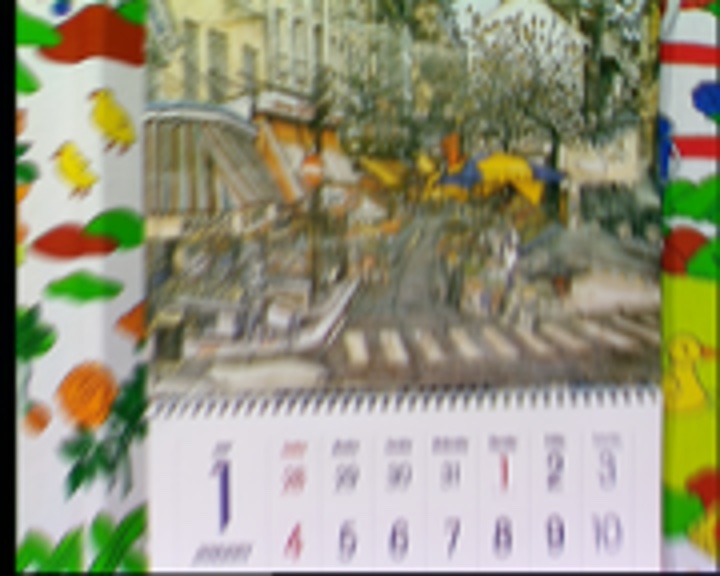}
							\\
							\tiny{Frame 30, Clip calendar, Vid4}~\cite{liu2013bayesianVid4} 
						\end{tabular}
					\end{adjustbox}
					\begin{adjustbox}{valign=t}
						\begin{tabular}{cccccc}
							\includegraphics[trim={470 0 190 500},clip,height=\h \textwidth, width=\w \textwidth]{\name bicubic.jpeg}  &
							\includegraphics[trim={470 0 190 500},clip,height=\h \textwidth, width=\w \textwidth]{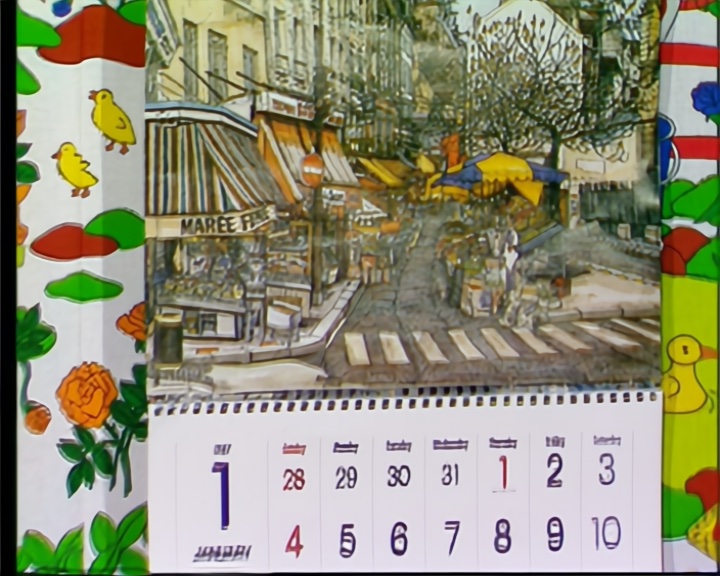} &
							\includegraphics[trim={470 0 190 500},clip,height=\h \textwidth, width=\w \textwidth]{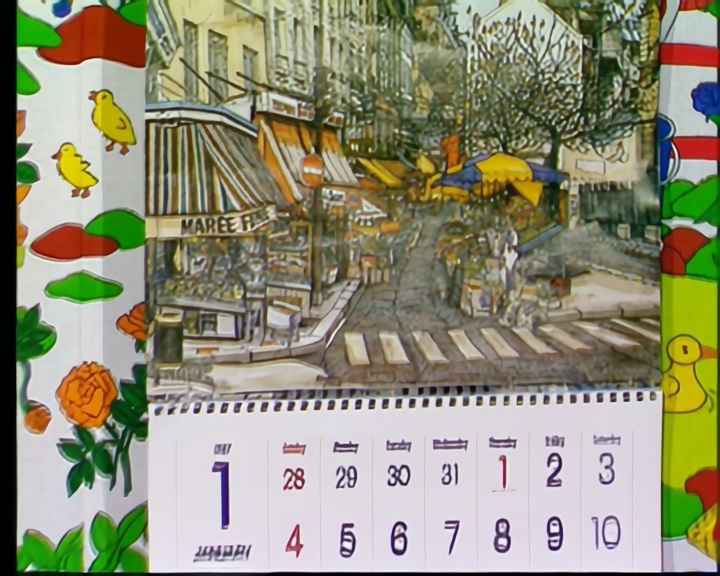} &
							\includegraphics[trim={470 0 190 500},clip,height=\h \textwidth, width=\w \textwidth]{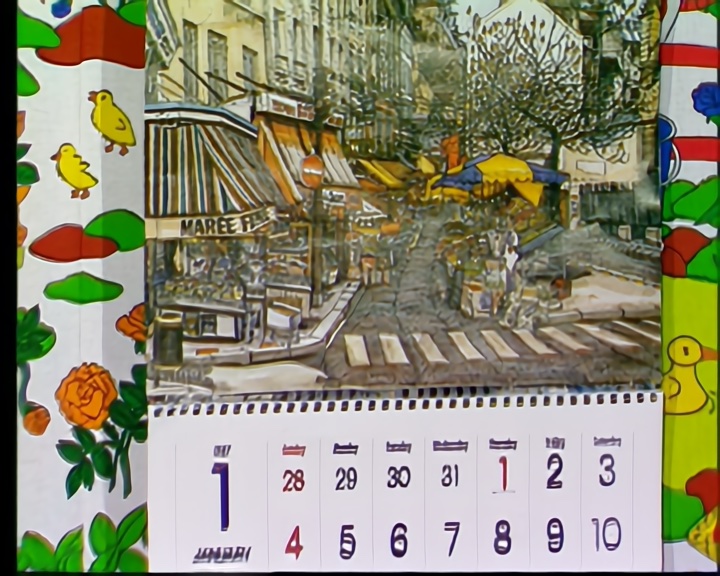} 
							\\
							\tiny{Bicubic} &
							\tiny{SloMo$\!$+BVSR} & \tiny{DAIN\!+BVSR} &
							\tiny{DAIN\!+CDVD\!+BVSR} 
							\\
							\includegraphics[trim={470 0 190 500},clip,height=\h \textwidth, width=\w \textwidth]{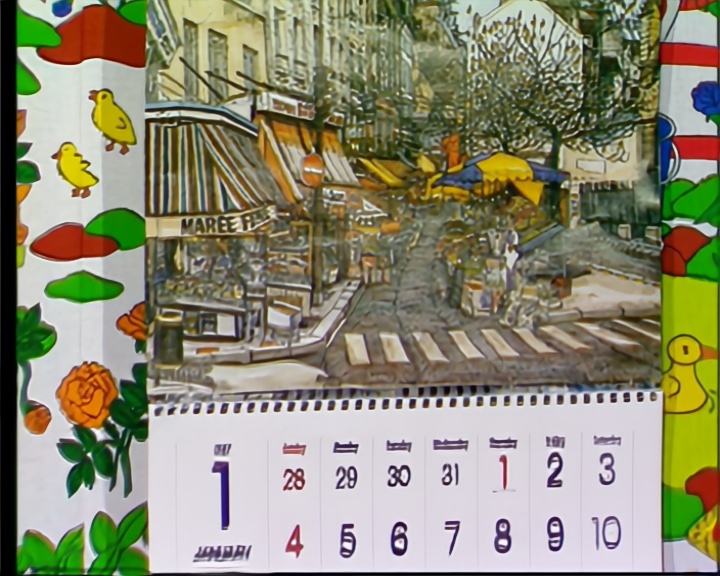} &
							\includegraphics[trim={470 0 190 500},clip,height=\h \textwidth, width=\w \textwidth]{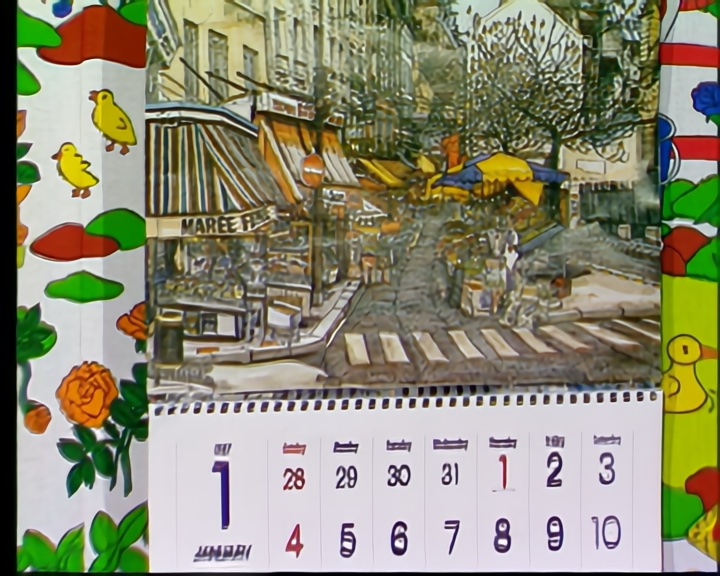} &
							\includegraphics[trim={470 0 190 505},clip,height=\h \textwidth, width=\w \textwidth]{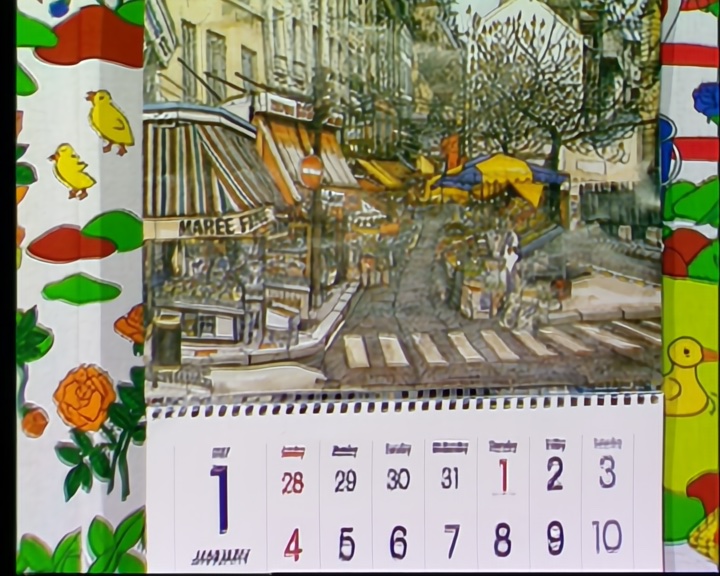}
							&		
							\includegraphics[trim={470 0 190 500},clip,height=\h \textwidth, width=\w \textwidth]{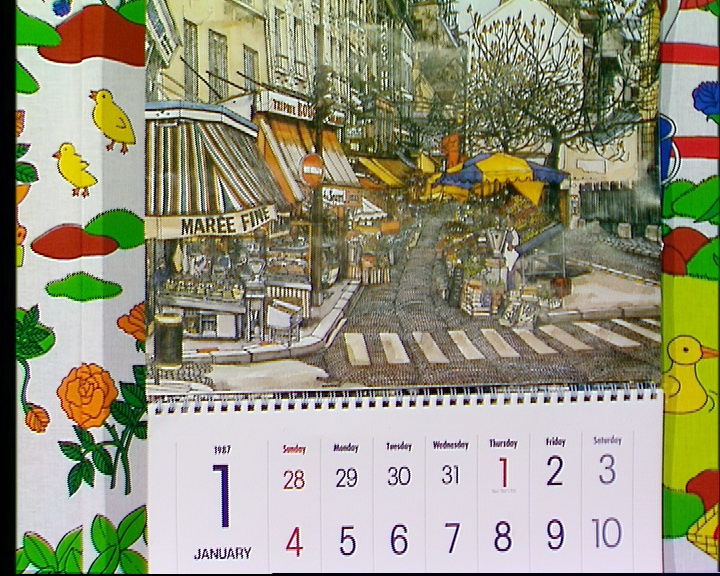} 
							\\ 
							\tiny{SloMo\!+EDVR$^{\dagger}$\!+BVSR}  &	\tiny{DAIN\!+EDVR$^{\dagger}$\!+BVSR}  &
							\tiny{\textbf{Ours}} &
							\tiny{GT}
							\\
						\end{tabular}
					\end{adjustbox}
				\end{tabular}
			\end{tabular}
		}
		\\ 
		\resizebox{1\linewidth}{!} {
			\begin{tabular}{cc}
				
				\renewcommand{\name}{figures/Vid4/00000016_}
				\renewcommand{\h}{0.12}
				\renewcommand{\w}{0.2}
				\begin{tabular}{cc}
					\begin{adjustbox}{valign=t}
						%\tiny
						\begin{tabular}{c}%left bottom right top
							\includegraphics[trim={130 10 0 20},clip, width=0.354\textwidth]{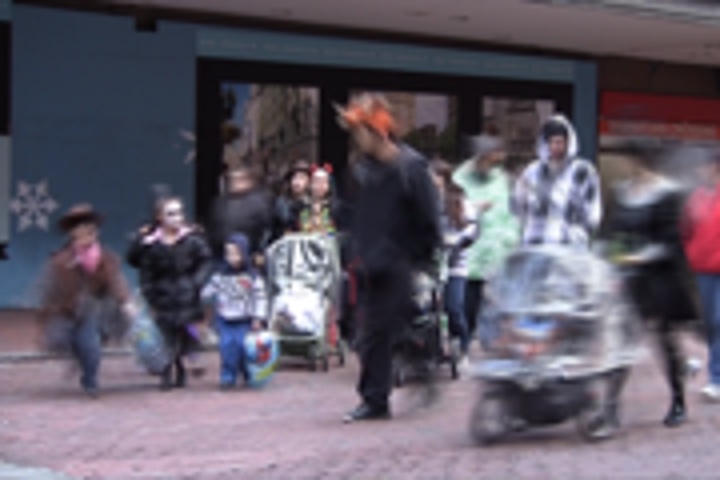}
							\\
							\tiny{Frame 16, Clip walk, Vid4}~\cite{liu2013bayesianVid4} 
						\end{tabular}
					\end{adjustbox}
					\begin{adjustbox}{valign=t}
						\begin{tabular}{cccccc}
							\includegraphics[trim={300 50 240 350},clip,height=\h \textwidth, width=\w \textwidth]{\name bicubic.jpeg}  &
							\includegraphics[trim={300 50 240 350},clip,height=\h \textwidth, width=\w \textwidth]{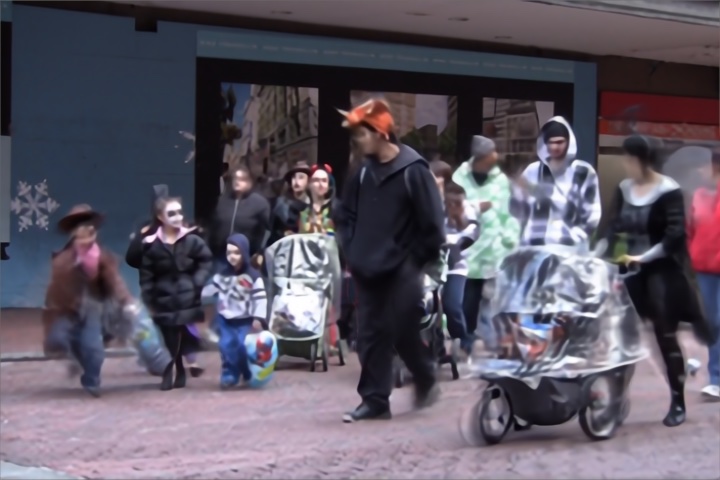} &
							\includegraphics[trim={300 50 240 350},clip,height=\h \textwidth, width=\w \textwidth]{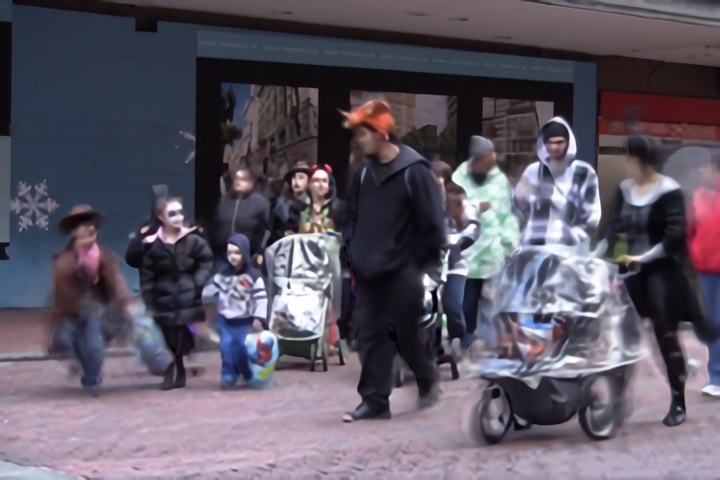} &
							\includegraphics[trim={300 50 240 350},clip,height=\h \textwidth, width=\w \textwidth]{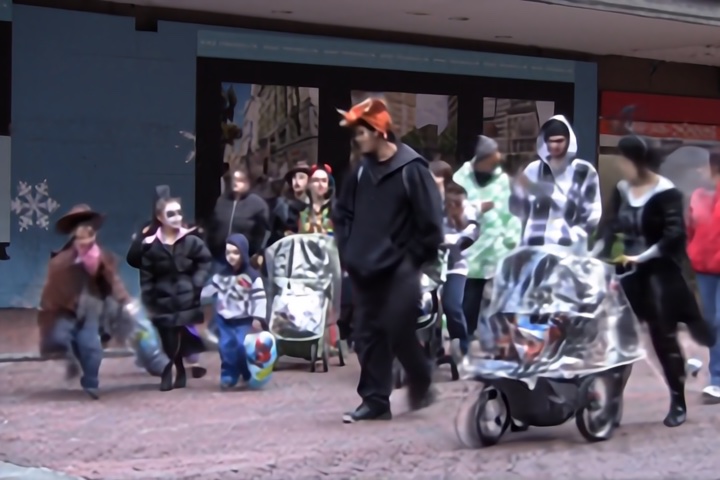} 
							\\
							\tiny{Bicubic} &
							\tiny{SloMo$\!$+BVSR} & \tiny{DAIN\!+BVSR} &
							\tiny{DAIN\!+CDVD\!+BVSR} 
							\\
							\includegraphics[trim={300 50 240 350},clip,height=\h \textwidth, width=\w \textwidth]{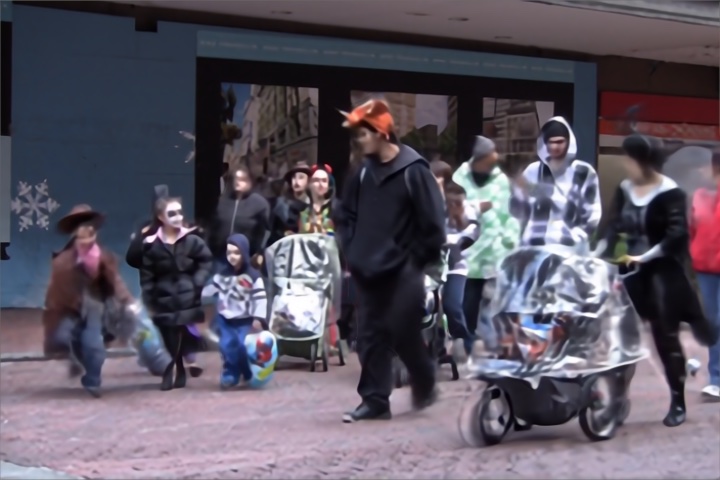}  &
							\includegraphics[trim={300 50 240 350},clip,height=\h \textwidth, width=\w \textwidth]{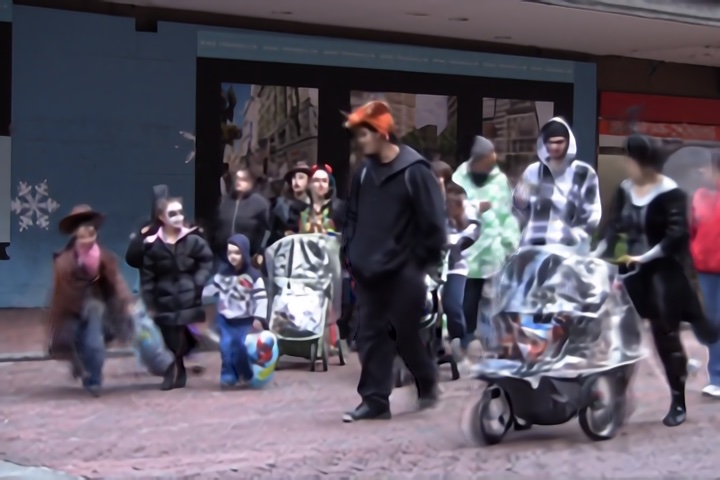} &
							\includegraphics[trim={300 50 240 350},clip,height=\h \textwidth, width=\w \textwidth]{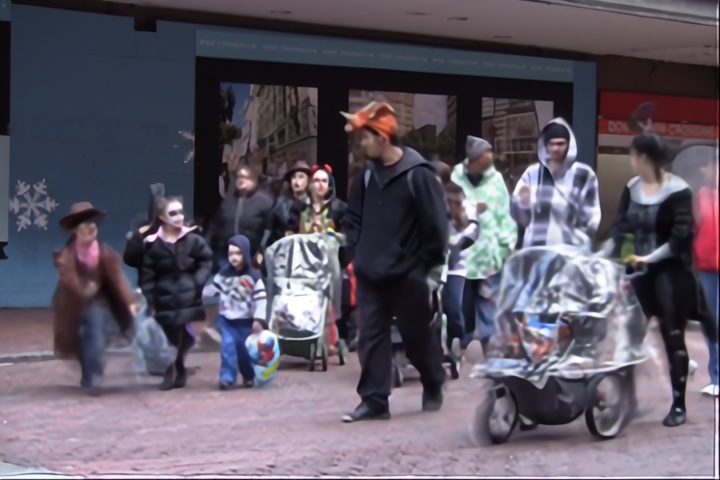}
							&		
							\includegraphics[trim={300 50 240 350},clip,height=\h \textwidth, width=\w \textwidth]{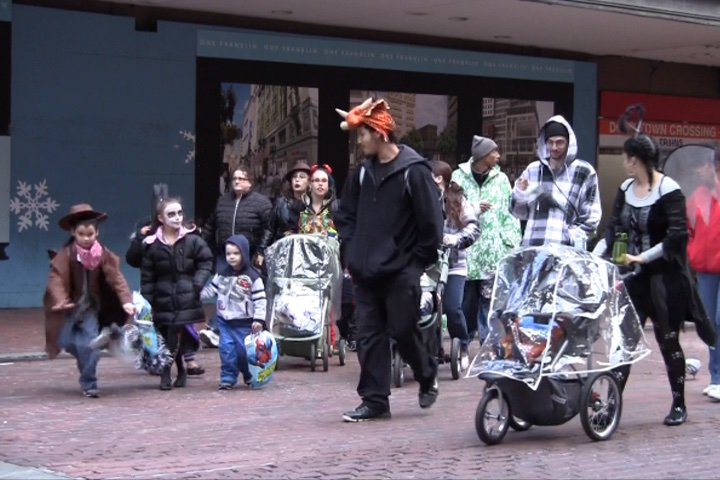} 
							\\ 
							\tiny{SloMo\!+EDVR$^{\dagger}$\!+BVSR}   &	\tiny{DAIN\!+EDVR$^{\dagger}$\!+BVSR}   &
							\tiny{\textbf{Ours}} &
							\tiny{GT}
							\\
						\end{tabular}
					\end{adjustbox}
				\end{tabular}
			\end{tabular}
		}
		\caption{Visual results of different methods on REDS4 and Vid4. Due to the space limitations of this figure, CDCD-TSP, SuperSloMo and BasicVSR++ are abbreviated as CDCD, SloMo and BVSR, respectively.} 
		\label{fig:sr_quali}
	\end{figure*}

	\noindent\textbf{Qualitative comparison.}
	Visual results of different methods are shown in Figure \ref{fig:sr_quali}. 
	In this figure, our proposed method achieves significant visual improvements over two-stage and three-stage methods. 
	This suggests that our one-stage framework is able to learn spatio-temporal information by exploring the natural intra-relatedness among the video interpolation, deblurring and super-resolution tasks.
	Compared with multi-stage methods, our method is able to synthesize high-quality HR video frames with clearer details and fewer blurring artifacts even for fast motion video sequences.
	Taking the second line~(Frame 040, Clip 011) as an example, our method is able to restore textures which are very close to GT although Clip\_011 has large motion.
	In contrast, multi-stage networks based on VFI methods (\ie Super-SloMo \cite{jiang2018superslomo} and DAIN \cite{bao2019dain}) tend to suffer from severe motion blur in the synthesized HR video, because it is difficult for the VFI methods to handle large motions in the videos. 
	More visual comparison results can be found in the supplementary materials.
	
	\begin{figure}[t]
        \setlength{\belowcaptionskip}{-0.3cm}
		\centering
		\includegraphics[width=1\linewidth,trim=0 40 0 20]{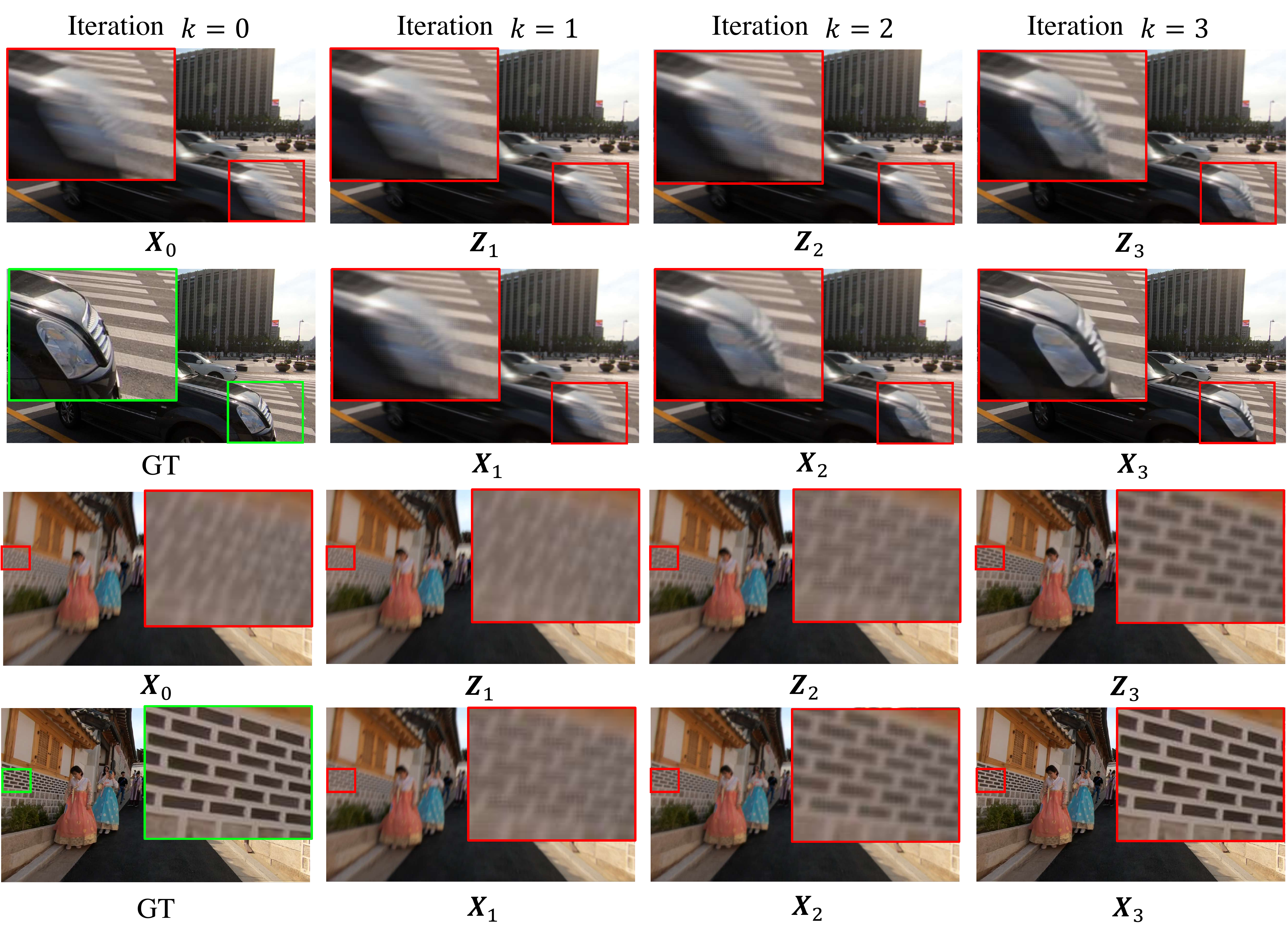}
		\caption{Visualization of FDT and RVE at different iterations on REDS4. 
		}
		\label{fig:vs_iter}
	\end{figure}

	\subsection{Visualization on Different Iterations}
	It is very interesting to investigate the synthesized outputs of the Fourier data transform (FDT) layer and the recurrent video enhancement (RVE) layer at different iterations. 
	The visualization results of our method at different iterations are provided in Fig. \ref{fig:vs_iter}. 
	As one can see, the quality of synthesized video frames continuously improves as the iteration number increases.
	This shows that, given an LSTR blurry video frame, the FDT and RVE layers are able to cooperatively and alternately deblur and recover high-frequency details of video frames.
	Specifically, the FDT layer eliminates blur kernel induced degradation and recovers tiny structures and fine textures, then the RVE layer restores the high-frequency information in videos.
	The visualization results demonstrate that our proposed DAVSR provides an interpretable way to understand STVSR.

	\subsection{Model Size and Inference Time}
	We investigate model sizes and runtime
	of different networks, and the results are shown in Fig. \ref{fig:para_psnr}. 
	To synthesize HSTR clear frames, the composed two-stage or three-stage methods can lead to very large model sizes for frame reconstruction. 
	In contrast, our one-stage model has much fewer parameters than the SOTA two-stage and three-stage networks. 
	For example, it is more than $3{\times}$ smaller than SuperSloMo+BasicVSR++$^*$. 
	With the help of the smaller model size, our model also achieves a fast inference time.
	More comparison results of model size and inference time can be found in the supplementary materials.

    \begin{wraptable}{r}{4cm}
    \setlength{\belowcaptionskip}{-0.85cm}
    \centering
        \caption{Ablation study on the FDT and EVD layers. Here we use PSNR as the evaluation metric.}
	   	\label{table:ablation}
		\resizebox{0.32\textwidth}{!}{
        \begin{tabular}{|c|c|c|c|}
        \hline
        FDT  & \XSolidBrush & \Checkmark   & \Checkmark \\ \hline 
        EVD  & \Checkmark   & \XSolidBrush & \Checkmark \\ \hline \hline
        ~PSNR~ & ~27.25~     & ~23.05~ & ~\bf{29.12}~ \\ \hline
        \end{tabular}
        }
    \end{wraptable}
    
	\subsection{Ablation Study}
\begin{figure}[t]
	\setlength{\belowcaptionskip}{-0.5cm}
	\setlength{\abovecaptionskip}{-0.01cm}
		\centering
		\includegraphics[width=1\linewidth]{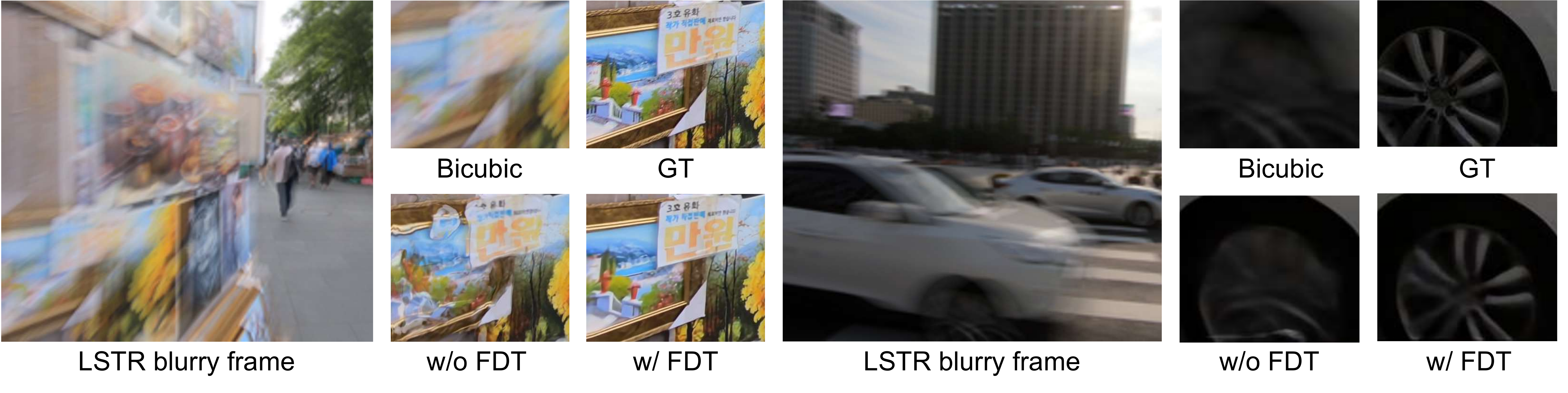}
		\caption{Visual results of the RDT layer on the REDS4 dataset.}
		\label{fig:ablation}
	\end{figure}

	We have already verified the superiority of our one-stage framework over two-stage and three-stage frameworks. 
	To further demonstrate the effectiveness of our network, we propose to conduct a comprehensive ablation study on the Fourier data transform (FDT) layer and the recurrent video enhancement (RVE) layer.
	As shown in Table \ref{table:ablation}, the model without the FDT or RVE layer is worse than original model.
	Next, we show the visual results of the ablation study in Fig. \ref{fig:ablation}.
	Here, we only conduct an ablation study on FDT because it contains no trainable parameters.
	If only the RVE layer is adopted without FDT, the synthesized SR video frames may suffer from blur artifacts. 
	On the other hand, if the data module is used without the RVE layer, the network cannot recover the high-frequency textures. This suggests that the FDT and EVD layers are complementary to each other and both are important to the proposed alternating optimization-based model. 
	
	\section{Conclusions}
	In this paper, we propose an alternating optimization for the practical space-time video super-resolution (STVSR) task by leveraging the model-based methods and learning-based methods.
	From an interpretable point of view, we first formulate the STVSR problem as two sub-problems related to the motion blur and motion aliasing.
	Specifically, we provide an analytical solution and propose a Fourier data transform layer to reduce the motion blur and motion aliasing for the first sub-problem. 
	Then, we propose a recurrent video enhancement layer in the second sub-problem to enhance the quality of the synthesized video.
	By the alternating optimization, our method is able to jointly handle the video interpolation, video deblurring and video super-resolution.
    Extensive experiments demonstrate that our framework achieves the state-of-the-art performance and it is more effective yet efficient than existing multi-stage networks.
	
    \paragraph{\textbf{\emph{Acknowledgements.}}}
	This work was partly supported by Huawei Fund and the ETH Zürich Fund (OK).
	
	\clearpage
	\bibliographystyle{splncs04}
	\bibliography{egbib}
\end{document}